\documentclass{article}

\PassOptionsToPackage{numbers,square}{natbib}
\usepackage[preprint]{neurips_2026}

\usepackage[utf8]{inputenc} 
\usepackage[T1]{fontenc}    
\usepackage{hyperref}       
\usepackage{url}            
\usepackage{booktabs}       
\usepackage{amsfonts}       
\usepackage{nicefrac}       
\usepackage{microtype}      
\usepackage{xcolor}         
\usepackage{graphicx}
\usepackage[ruled,vlined,linesnumbered]{algorithm2e}
\usepackage{amsmath}
\usepackage{multirow} 
\usepackage{subcaption}
\usepackage{wrapfig}
\usepackage[table]{xcolor}
\definecolor{lightblue}{RGB}{235,245,255}
\definecolor{lightred}{RGB}{255,230,230}
\definecolor{darkblue}{RGB}{0,0,139}
\definecolor{myyellow}{RGB}{220,180,0}
\title{Bridging Local Observation and Global Simulation in Closed-Loop Traffic Modeling}

\author{%
    Ziyan Wang$^{1,2}$\footnotemark[1]\quad\quad
    Tan Xiang$^2$\footnotemark[1]\quad\quad
    Peng Chen$^2$\quad\quad
    Xintao Yan$^1$\footnotemark[2]\\
$^1$ The University of Hong Kong \quad$^2$ Beihang University\\
}

\begin{document}

\maketitle

\renewcommand{\thefootnote}{\fnsymbol{footnote}}
\footnotetext[1]{Equal contribution.} \footnotetext[2]{Corresponding author. (Email: \texttt{xintaoy@hku.hk})}

\begin{abstract}

A local-to-global context mismatch arises when autoregressive traffic simulators trained on ego-centric driving logs are deployed in globally observable closed-loop environments. In such logs, the ego vehicle has rich local observations, while surrounding agents are only partially observed due to perception limits and occlusions. As a result, simulators may learn incomplete context--action mappings that remain hidden in log-based training but emerge during closed-loop rollouts, leading to unrealistic behaviors such as abnormal stops, unsafe interactions, and rule violations. We propose \textbf{CRAFT}, a \textbf{C}ontextual p\textbf{R}eference \textbf{A}lignment \textbf{F}ramework for \textbf{T}raffic Simulation, to mitigate this mismatch via self-supervised failure discovery and preference-guided test-time alignment. CRAFT treats the base simulator as a globally observable sandbox, generating diverse what-if rollouts from logged initial states to expose context-induced failures. These failures are grounded with human-aligned driving priors and converted into preference supervision for training a Contextual Preference Evaluator (CPE). At inference time, CPE acts as a plug-in alignment module that scores candidate actions under complete scene context and reweights autoregressive decoding toward globally coherent behaviors. CRAFT mitigates this local-to-global contextual bias, reducing collisions by 31.2\% and traffic violations by 33.2\% without retraining the base simulator. The official implementation is available
at  \tt{\href{https://ashorizon.github.io/CRAFT-site/}{\tt https://ashorizon.github.io/CRAFT-site/}}.

\end{abstract}

\section{Introduction}
\label{sec:intro}
Realistic traffic simulation is a critical generative data engine for autonomous driving, enabling scalable training and evaluation in complex multi-agent environments~\cite{feng2023dense,yan2023learning}. Recent advances in generative modeling have led to the widespread adoption of autoregressive architectures~\cite{seff2023motionlm, philion2023trajeglish}, which formulate traffic simulation as sequential trajectory generation. Most existing approaches~\cite{suo2021trafficsim, yan2023learning, zhang2023trafficbots, wu2024smart} train simulators on large-scale ego-centric driving logs under an imitation learning paradigm, learning context–trajectory mappings by maximizing the likelihood of logged behaviors. This paradigm has substantially advanced the realism and fidelity of microscopic traffic simulation, enabling realistic multi-agent interactions and achieving strong performance on widely used benchmarks~\cite{montali2023waymo}.

\begin{figure}
  \centering
  \includegraphics[width=\linewidth]{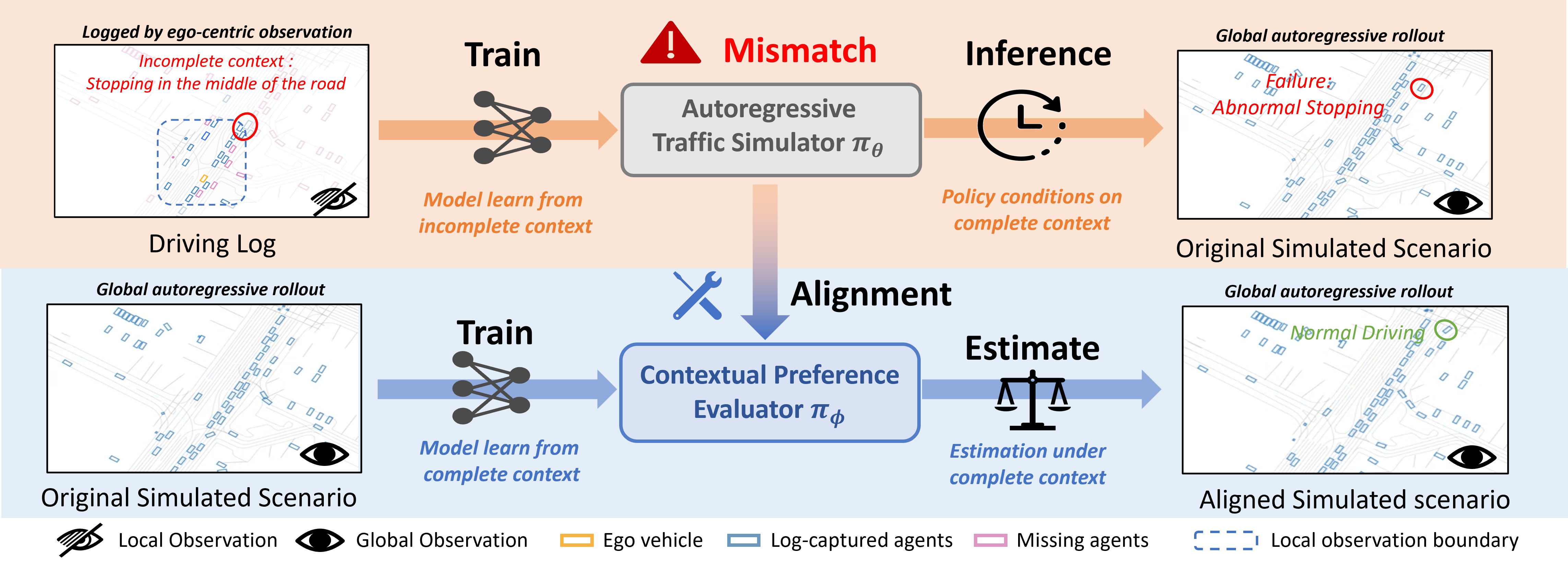}
    \caption{
    \textbf{Motivation of CRAFT.}
Incomplete ego-centric logs can induce flawed context--action mappings, leading to abnormal behaviors in global autoregressive rollouts. CRAFT mitigates this mismatch by learning complete-context preferences and guiding simulation toward rational behaviors.
    }
    \label{fig:motivation}
    \vspace{-1em}
\end{figure}

Despite these advances, a fundamental question remains largely unexplored: \emph{can policies learned from locally observed driving logs remain behaviorally rational when executed in globally contextualized closed-loop simulation?} Figure~\ref{fig:motivation} illustrates why this assumption may fail. In real-world data collection, a leading vehicle may brake in response to a downstream obstacle. However, if the obstacle lies outside the perception range of the logging vehicle or is occluded by other agents, the recorded log captures the braking behavior while omitting the context that justifies it. From the perspective of the observed data, this valid maneuver appears as an unexplained stop. When such partial observations are treated as complete supervision, the learned model may associate braking with incomplete context, reproducing the action even when the true cause is absent \cite{geirhos2020shortcut}. This leads to ambiguous context–action mappings and unrealistic behaviors during closed-loop rollout \cite{si2023measuring, de2019causal}.

This phenomenon reflects a broader local-to-global mismatch in traffic simulation \cite{meng2026infrastructure}. Most existing simulators are trained on large-scale driving logs collected from ego-centric vehicle platforms, such as Waymo~\cite{ettinger2021large}, nuPlan~\cite{caesar2021nuplan}, and Argoverse~\cite{wilson2023argoverse}, where perception boundaries, occlusions, and sensor viewpoints are inherent. As a result, these logs provide rich local driving priors but limited evidence about how a model should behave under full scene context during long-horizon rollouts. The key challenge is therefore how to mitigate abnormal behaviors induced by incomplete context-action learning, especially when these behaviors are not directly observed in the original logs but emerge during closed-loop generation \cite{codevilla2019exploring}.

To address this challenge, we propose to shift from passive imitation to active alignment. Instead of viewing simulation solely from the perspective of the biased autoregressive generator, we introduce a complementary \emph{Contextual Preference Evaluator} (CPE) that assesses the rationality of generated behaviors under complete scene context. We use the base simulator itself as a globally observable sandbox: starting from logged initial states, it generates multiple what-if rollouts that reveal how biases learned from incomplete observations manifest as closed-loop failures. We further ground these self-exposed failures with human-aligned driving priors and transform them into preference supervision. The CPE is then trained with contrastive preference learning to distinguish globally coherent behaviors from locally plausible but contextually inconsistent ones. During inference, CPE acts as a plug-in test-time alignment module that reweights candidate actions from the frozen simulator, guiding autoregressive decoding toward behaviors that are more consistent with the full traffic scene. This converts traffic simulation from passive next-token sampling into preference-guided decision making, improving closed-loop behavioral rationality.

\paragraph{Contributions.} \textbf{(1)} We identify a fundamental mismatch between local observations and global simulation requirements, showing that imitation learning on ego-centric logs leads to incomplete context–action associations that degrade closed-loop behavior. \textbf{(2)} We propose a decoupled \textit{Contextual Preference Evaluator} (CPE) that learns complete-context behavioral preferences from simulator-induced failures grounded by human-aligned driving priors. \textbf{(3)} We deploy CPE as a plug-in test-time guidance module for autoregressive decoding, improving behavioral rationality across simulators while preserving competitive realism. Extensive experiments demonstrate that our approach bridges local-to-global mismatch, achieving superior performance in closed-loop traffic modeling.

\section{Related Work}

\paragraph{Generative Traffic Models from Driving Logs.}
Modeling multi-agent traffic behavior is central to autonomous driving simulation. Early predictive models, such as QCNet~\cite{zhou2023query} and Multiverse Transformer~\cite{wang2023multiverse}, regress future trajectories from historical states, but often struggle to capture joint dependencies required for closed-loop interaction. Recent generative models learn richer behavior distributions: diffusion-based methods, including VBD~\cite{huang2024versatile}, CTG~\cite{zhong2023guided}, and NEXUS~\cite{zhou2025decoupled}, generate diverse futures through iterative denoising, while autoregressive models such as Trajeglish~\cite{philion2023trajeglish}, BehaviorGPT~\cite{zhou2024behaviorgpt}, and SMART~\cite{wu2024smart} discretize driving states into motion tokens for next-token prediction. Benchmarks such as WOSAC~\cite{montali2023waymo} further promote closed-loop evaluation. However, most simulators are still trained from ego-centric logs, whose partial observations may induce spurious context-action associations and poor closed-loop generalization~\cite{de2019causal,codevilla2019exploring,vinitsky2022nocturne,nguyen2025lead}.

\paragraph{Closed-Loop Refinement of Traffic Simulators.}
Recent work refines pretrained driving models after imitation learning to reduce closed-loop errors. Supervised fine-tuning methods, such as UniMM~\cite{lin2025revisit} and CAT-K~\cite{zhang2025closed}, generate rollouts, select high-quality trajectories, and continue training on the selected samples. Reinforcement learning fine-tuning methods, including SMART-R1~\cite{pei2025advancing}, R1Sim~\cite{wang2026learning}, and LLM2AD~\cite{wang2025llm}, optimize policies with rewards that encourage expert-like behavior or penalize violations. These methods show the importance of adapting simulators to their own rollout distributions, but they still rely on training-time optimization of high-dimensional generators, often with carefully designed reward shaping or simulation pipelines. Such optimization can suffer from sparse or delayed feedback, credit-assignment difficulty, and stability issues~\cite{gao2026rad}. We instead study a complementary test-time correction strategy without updating the base simulator.

\paragraph{Preference Evaluation and Test-Time Guidance.}
Evaluation-guided generation has been widely used to improve driving reliability \cite{zhao2026bridgesim}. Rule-based evaluator provide interpretable criteria for collisions, offroad behaviors, traffic-rule violations, and unfeasible motions, and are commonly used to filter, rank, or evaluate generated trajectories~\cite{dauner2023parting}. Beyond handcrafted criteria, learned and generation-time evaluators have also been explored. Gen-Drive~\cite{huang2025gen} trains a scene evaluator for reward modeling and policy refinement, while SaFeR~\cite{cui2026safer} performs feasibility-constrained token resampling to generate safety-critical but plausible scenarios. Unlike methods that use evaluation mainly for post-hoc ranking, reward modeling, or safety-critical generation, our work incorporates preference evaluation directly into test-time decoding for closed-loop traffic simulation, guiding generated behaviors toward greater realism and rationality.

\section{Problem Formulation}

\textbf{Autoregressive traffic simulation.}
We consider closed-loop traffic simulation initialized from a logged driving scene state. 
Let $\mathbf{C}_t$ denote the \emph{dynamic scene state} at time $t$, which includes all agent states and traffic signal states that evolve over time. 
Let $\mathbf{X}_M$ denote the \emph{static map context} of the environment, including road geometry and lane topology. Given an initial state $\mathbf{C}_0$ and map $\mathbf{X}_M$, a pretrained autoregressive simulator $\pi_\theta$ (with parameters $\theta$) models the evolution of the scene by sequentially generating joint actions for all agents. 
At each step $t$, the simulator outputs a distribution $P_\theta$ over joint action tokens, from which a joint action $\mathbf{a}^{t+1}$ is sampled:

\begin{equation}
    P_\theta(\cdot \mid \mathbf{C}_t,\mathbf{X}_M)
    =
    \pi_\theta(\mathbf{C}_t,\mathbf{X}_M),
    \qquad
    \mathbf{a}^{t+1}
    \sim
    P_\theta(\cdot \mid \mathbf{C}_t,\mathbf{X}_M),
    \qquad
    \mathbf{C}_{t+1}
    =
    \operatorname{Step}(\mathbf{C}_t,\mathbf{a}^{t+1}),
\end{equation}
where $\operatorname{Step}(\cdot)$ is the deterministic simulator transition function. The resulting autoregressive rollout over a horizon $T_f$ is given by
\begin{equation}
    P_\theta(\mathbf{a}^{1:T_f} \mid \mathbf{C}_0, \mathbf{X}_M)
    =
    \prod_{t=0}^{T_f-1}
    \pi_\theta(\mathbf{a}^{t+1} \mid \mathbf{C}_t, \mathbf{X}_M),
    \quad
    \mathbf{C}_{t+1} = \operatorname{Step}(\mathbf{C}_t, \mathbf{a}^{t+1}).
    \label{eq:rollout}
\end{equation}

\textbf{Preference-guided alignment objective.} We keep the base simulator $\pi_\theta$ fixed and modifies only its test-time decoding distribution by introducing an evaluator $\pi_\phi$ to assess the scenes induced by candidate actions before execution. Given a candidate action $\mathbf{a}^{t+1}$ sampled from the base distribution $P_\theta$, a one-step lookahead scene is first constructed and preference score is estimated by the evaluator:
\begin{equation}
    \mathbf{a}^{t+1}
    \sim
    P_\theta(\cdot \mid \mathbf{C}_t,\mathbf{X}_M),
    \qquad
    \tilde{\mathbf{C}}_{t+1}
    =
    \operatorname{Step}(\mathbf{C}_t,\mathbf{a}^{t+1}),
    \qquad
    \mathbf{R}_{t}
    =
    \pi_\phi(\tilde{\mathbf{C}}_{t+1},\mathbf{X}_M).
\end{equation}
Here $\mathbf{R}_{t}$ measures whether the candidate action leads to a new scene under the current context. The policy updates the original decoding distribution using these scene-level preference scores:
\begin{equation}
    P_{\theta,\phi}(\mathbf{a}^{t+1}\mid \mathbf{C}_t,\mathbf{X}_M)
    =
    \operatorname{Calibrate}
    \left(
    P_\theta(\mathbf{a}^{t+1}\mid \mathbf{C}_t,\mathbf{X}_M);
    \mathbf{R}_{t}
    \right).
\end{equation}

\section{Method}

\subsection{Grouped complete-context what-if simulation}
\label{sec:grouped_complete_context_what_if_simulation}
\begin{figure}
  \centering
  \includegraphics[width=.78\linewidth]{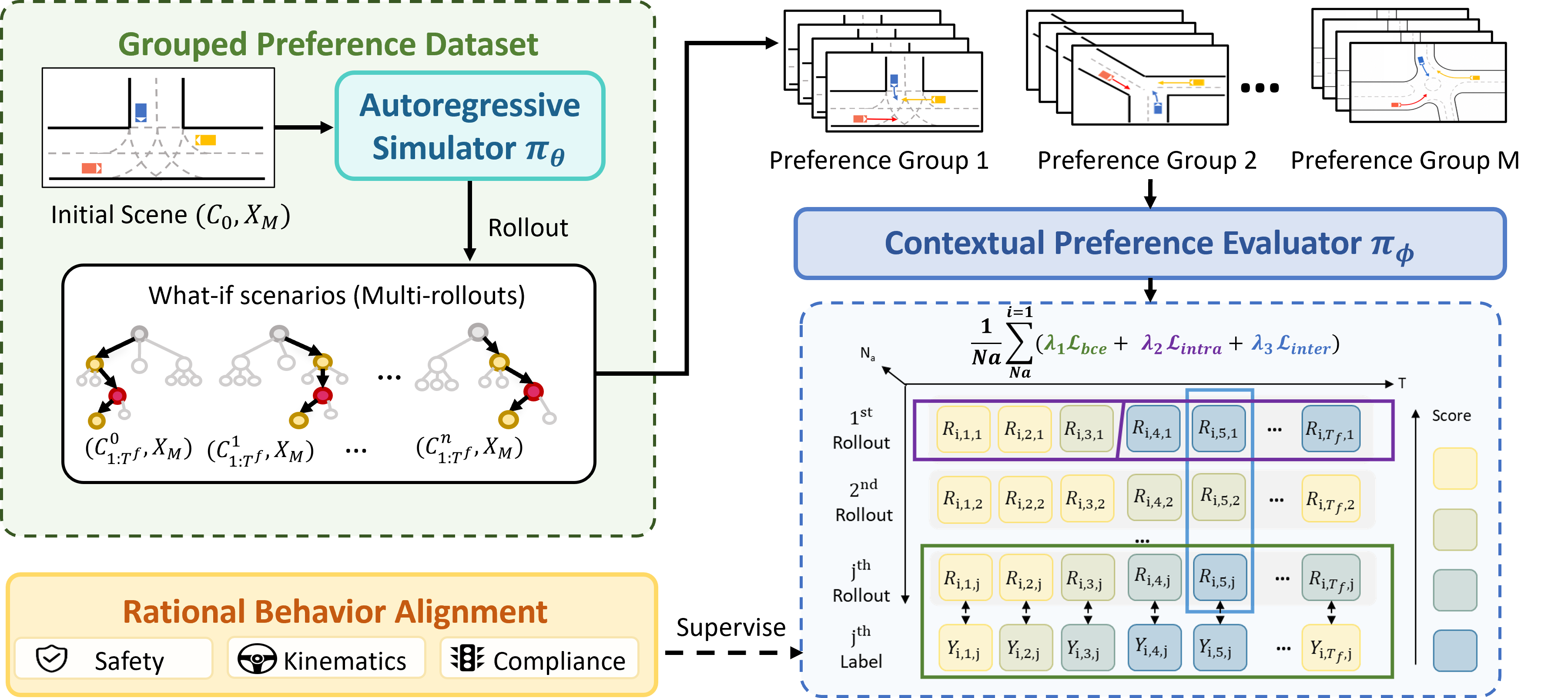}
\caption{
\textbf{The training pipeline of CRAFT.} 
Starting from logged initial scenes, a frozen autoregressive simulator generates complete-context what-if rollouts, which are annotated by rule-based evaluators and organized into grouped preference datasets.
The CPE is trained on grouped rollouts to predict dense preference scores with token-level, intra-trajectory, and inter-rollout supervision.
}

    \label{fig:train}
\end{figure}

As shown in Fig.\ref{fig:train}, given an original logged dataset $\mathcal{D}_{\mathrm{log}}=\{(\mathbf{C}_0^m,\mathbf{X}_M^m)\}_{m=1}^{\mathcal{M}}$, we employ the pretrained autoregressive simulator $\pi_\theta$ as a what-if simulation generator. For each initial scene $(\mathbf{C}_0^m,\mathbf{X}_M^m)$, the base simulator performs $N$ stochastic rollouts with diverse random seeds followed by equation \ref{eq:rollout}:
\begin{equation}
    \hat{\mathbf{C}}_{1:T_f}^{(m,j)}
    =
    \operatorname{Rollout}
    \left(
    \mathbf{C}_0^m,\mathbf{X}_M^m;\epsilon_j
    \right),
    \qquad
    j=1,\ldots,N ,
\end{equation}
where $\epsilon_j$ denotes the random sampling seed. Through multimodal sampling and autoregressive scene updates, these rollouts evolve into diverse what-if simulations where all represented agents interact under the same map context.

For each rollout, we derive agent-time preference supervision from human-aligned driving priors, including safety, rule compliance, and kinematic feasibility. Each token is assigned a binary label $y_{i,t}\in\{0,1\}$ from the auto-labeling evaluation, where positive labels indicate reasonable behaviors and negative labels indicate abnormal behaviors:
\begin{equation}
    \mathbf{Y}_{1:T_f}^{(m,j)}
    =
    \mathcal{A}
    \left(
    \hat{\mathbf{C}}_{1:T_f}^{(m,j)},\mathbf{X}_M^m
    \right),
    \qquad
    \mathbf{Y}_{1:T_f}^{(m,j)}\in\{0,1\}^{N_a\times T_f}.
\end{equation}
The resulting preference dataset is organized as grouped alternative futures:
\begin{equation}
    \mathcal{D}_{\mathrm{pref}}
    =
    \{\mathcal{G}_m\}_{m=1}^{\mathcal{M}},
    \qquad
    \mathcal{G}_m
    =
    \left\{
    \left(
    \hat{\mathbf{C}}_{1:T_f}^{(m,j)},
    \mathbf{Y}_{1:T_f}^{(m,j)}
    \right)
    \right\}_{j=1}^{N}.
\end{equation}
This grouped construction provides comparable rollouts from the same logged scene, supporting both token-level supervision and inter-rollout preference learning. More details on the auto-labeling criteria and the construction of the grouped preference dataset are provided in the Appendix \ref{sec:C}.

\begin{figure}
  \centering
  \includegraphics[width=.9\linewidth]{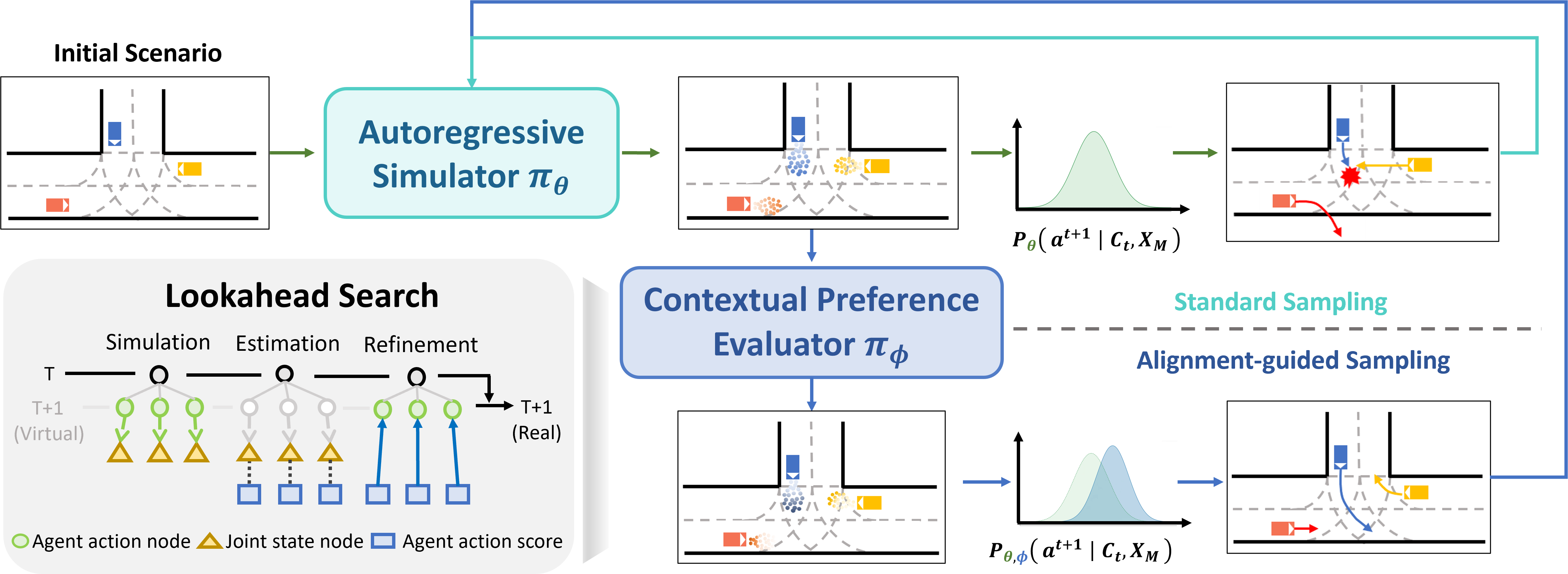}
    \caption{\textbf{The inference pipeline of CRAFT.} During inference, CPE evaluates behaviors via a lookahead search algorithm, computing the posterior probabilities before sampling. 
  }
    \label{fig:pipeline_2}
\end{figure}
\subsection{Complete-Context Preference Learning}
\label{sec:complete_context_preference_learning}

Leveraging $\mathcal{D}_{\mathrm{pref}}$, we train a Contextual Preference Evaluator (CPE) to estimate the behavioral rationality of motion tokens in Fig.~\ref{fig:train}. Given a grouped what-if rollout $\hat{\mathbf{C}}^{m,j}_{0:T_f}$ and map context $\mathbf{X}^{m}_{M}$, CPE estimates a dense agent-time preference matrix:
\begin{equation}
    \mathbf{R}^{m,j}
    =
    \pi_{\phi}
    \left(
    \hat{\mathbf{C}}^{m,j}_{0:T_f},
    \mathbf{X}^{m}_{M}
    \right),
    \qquad
    \mathbf{R}^{m,j}\in[0,1]^{N_a\times (T_f)} ,
\end{equation}
where $R_{i,t}^{m,j}$ estimates whether agent $i$'s behavior at step $t$ is reasonable under the complete simulated scene. CPE uses the frozen scene encoder of the base simulator and a temporal convolutional preference decoder, which processes each agent's temporal context and estimates the score matrix in a single parallelized forward pass.

CPE is trained with absolute token-level supervision and relative preference constraints. We formulate the weighted Binary Cross Entropy loss \cite{shannon1948mathematical} as
\begin{equation}
    \mathcal{L}_{bce} = - \frac{1}{N_a (T_f)} \sum_{i=1}^{N_a} \sum_{t=1}^{T+T_f} \left[ y_{i,t} \log(R_{i,t}) + (1 - y_{i,t}) \log(1 - R_{i,t}) \right],
\end{equation}

Beyond token-level labels, contrastive preference learning provides relative supervision that encourages CPE to distinguish subtle differences in behavioral quality. We introduce two complementary preference losses: an intra-trajectory loss that exploits temporal transitions from normal behavior to failure within a rollout, and an inter-rollout loss that leverages the grouped structure of $\mathcal{D}_{\mathrm{pref}}$ to compare alternative futures generated from the same initial scene.

For intra-trajectory alignment, $\mathcal{P}_{intra}$ contains agents that encounter a failure event within a single rollout. For each $i\in\mathcal{P}_{intra}$, a normal segment $T^+$ is preferred over a subsequent failure segment $T^-$. We apply a margin ranking loss \cite{bordes2013translating}:
\begin{equation}
    \mathcal{L}_{intra} = \frac{1}{|\mathcal{P}_{intra}|} \sum_{i \in \mathcal{P}_{intra}} \max \left( 0, \bar{R}_{i,T^-} - \bar{R}_{i,T^+} + m \right).
\end{equation}

For inter-rollout alignment, $\mathcal{P}_{inter}$ contains token-level pairs $(i,t,j^+,j^-)$ from different rollouts in the same group. Here, $j^+$ and $j^-$ denote preferred and less preferred futures for the same agent and time step. We optimize the Bradley--Terry objective \cite{bradley1952rank}:
\begin{equation}
    \mathcal{L}_{inter} = - \frac{1}{|\mathcal{P}_{inter}|} \sum_{(i, t, j^+, j^-) \in \mathcal{P}_{inter}} \log \frac{\exp \left( R_{i,t}^{(j^+)} \right)}{\exp \left( R_{i,t}^{(j^+)} \right) + \exp \left( R_{i,t}^{(j^-)} \right)}.
\end{equation}

The overall training objective is
\begin{equation}
    \mathcal{L}_{CPE} = \lambda_{1} \mathcal{L}_{bce} + \lambda_{2} \mathcal{L}_{intra} + \lambda_{3} \mathcal{L}_{inter},
\end{equation}
where $\lambda_{1}$, $\lambda_{2}$, and $\lambda_{3}$ balance absolute token-level supervision and relative preference constraints.

\subsection{Test-Time Preference Alignment}
\label{sec:test_time_guidance}

After training, CPE is used to guide autoregressive sampling by evaluating the next-step candidates proposed as shown in Fig.\ref{fig:pipeline_2}. At step $t$, each agent receives $K$ candidate action tokens from the original decoding distribution. Each candidate joint action is first applied to the current scene to construct a one-step lookahead scene, whose behavioral rationality is scored by CPE. This design aims to calibrate the simulator's next-token distribution before action execution. Ideally, evaluating every possible multi-agent action combination would provide the preference score for the full joint action space. However, for $N_a$ agents with $K$ candidates each, exhaustive evaluation requires
$K^{N_a}$. CPE forward passes at each step, incurring high computational cost in dense traffic scenes.

Instead of exhaustively evaluating the intractable $K^{N_a}$, joint action space, we adopt an Independent Marginal Approximation strategy. By drawing $K$ parallel joint action samples from the base simulator's prior, we effectively construct $K$ independent Monte Carlo rollout scenes. For any individual agent, the actions of all other agents in the $k$-th scene act as a highly probable stochastic background. This transforms the exponential joint evaluation into a linear $O(K)$ marginal evaluation, estimating the expected contextual preference score of an action under plausible background traffic. This decoupled evaluation is empirically sound due to the kinematic inertia within a single autoregressive step (e.g., 0.1s-0.5s). In such a short lookahead horizon, severe anomalies are primarily determined by the agent's own marginal action rather than the exact micro-interactions of distant agents. Therefore, evaluating actions against a sampled plausible background provides sufficient localized evidence to reject explicitly irrational choices. Algorithm~\ref{alg:guidance} summarizes the procedure.

\begin{algorithm}
\caption{Inference Sampling Guidance}
\label{alg:guidance}
\SetKwInOut{KwIn}{Input}
\KwIn{Initial scene context $\mathbf{C}_{0}$; Base simulator $\pi_{\theta}$; Contextual Preference Evaluator $\pi_{\phi}$; Scenario horizon $T_f$; Candidates $K$; Guidance scale $\beta$.}

\For{$t = 1,\ldots,T_f$}{
    Extract Top-$K$ candidate action pool $\mathbf{A}_{t} \in \mathbb{R}^{N_a \times K}$ and priors $\mathbf{P}^{\text{prior}}_{t}$ from $\pi_{\theta}(\cdot | \mathbf{C}_{t-1})$\;

    \For{candidate index $k = 1,\ldots,K$}{
        Generate the lookahead scenario:
        $\tilde{\mathbf{C}}_{t}^{(k)} \leftarrow \texttt{Step}(\mathbf{C}_{t-1}, \tilde{\mathbf{a}}_{t}^{(k)})$. \;

        Evaluate alignment scores:
        $\mathbf{R}_{t}^{(k)} \leftarrow \pi_{\phi}(\tilde{\mathbf{C}}_{t}^{(k)})$, where $\mathbf{R}_{t}^{(k)} \in \mathbb{R}^{N_a}$. \;
        
        Update probabilities:
        $\mathbf{P}^{\text{unnorm}}_{:,t,k}\leftarrow \mathbf{P}^{\text{prior}}_{:,t,k} \cdot \exp(\beta \cdot \mathbf{R}_{t}^{(k)})$. \;
    }

    Normalize probabilities across candidates:
    $\mathbf{P}^{\text{post}}_{i,t,k} \leftarrow \frac{P^{\text{unnorm}}_{i,t,k}}{\sum_{j=1}^{K} P^{\text{unnorm}}_{i,t,j}}$. \;

    Sample $a_{i,t}\sim\texttt{Categorical}(\mathbf{P}^{\text{post}}_{i,t,:})$ for all agents $i$\;

    Execute the actual step update:
    $\mathbf{C}_{t} \leftarrow \texttt{Step}(\mathbf{C}_{t-1}, \{a_{i,t}\}_{i=1}^{N_a})$. \;
}

\Return{$\mathbf{C}_{1:T_f}$}.
\end{algorithm}

\section{Experiment}
\label{sec:experiment}
\subsection{Experimental Setups}
\textbf{Datasets and base simulator.} We employ the Waymo Open Motion Dataset (WOMD) \cite{ettinger2021large} as the training and validation dataset for the experiments. The WOMD comprises more than 480k multi-agent interactive scenarios, each with a duration of 9 seconds and sampled at 10 Hz, which is suitable for the training and validation of traffic simulation models. For the base autoregressive traffic simulator, we utilize the CAT-K model\cite{zhang2025closed} pre-trained on the WOMD.

\textbf{Metrics.}
We evaluate traffic simulation under two complementary protocols. 
First, we report the official WOSAC metrics~\cite{montali2023waymo} to ensure comparability with the standard benchmark. This protocol mainly measures log-similarity and realism on benchmark-defined scored agents, and also provides safety-related rates such as collision, offroad, and traffic-light violations. 
Second, we report all-agent behavioral metrics following prior simulation studies~\cite{lu2024scenecontrol,rowe2025ctrl}. This protocol evaluates all moving agents in the generated scene, focusing on distributional realism and behavioral rationality, including collision, offroad, and traffic-rule violation rates at both agent and scene levels. 
The two protocols are complementary: WOSAC emphasizes standard realism comparison, while the all-agent protocol better reflects whether a closed-loop simulator produces safe and compliant behaviors for the entire traffic scene. 
More details are provided in the Appendix \ref{sec:B}.

\textbf{Baselines.}
To comprehensively evaluate our framework, we compare it with representative baselines from two categories. 
(1) \textit{Pretrained imitation learning models}: GUMP~\cite{hu2024solving} is a generative motion model for multi-task motion generation, and SMART~\cite{wu2024smart} is an autoregressive multi-agent simulator trained via next-token behavior cloning. 
(2) \textit{Post-training refinement models}: CAT-K~\cite{zhang2025closed} refines tokenized traffic models through closed-loop supervised fine-tuning, while R1Sim~\cite{wang2026learning} applies reinforcement learning fine-tuning with handcrafted rewards to optimize generated trajectories. 
(3) \textit{Auto-labeler}: We additionally compare with a direct rule-based guidance baseline, where the same human-aligned criteria used for auto-labeling, are applied online to score candidate actions. 

\begin{table}[t]
\centering
\caption{Simulation performance comparison under\textbf{ WOSAC metrics}. Col., Off., and Tra. denote collision rate, offroad rate, and traffic violation rate, respectively.}
\resizebox{\linewidth}{!}{
\begin{tabular}{l l c c c c c c c c c}
\toprule
\multirow{2}{*}{Base Model}
& \multirow{2}{*}{Strategy}
& \multicolumn{4}{c}{WOSAC Metrics $\uparrow$}
& \multicolumn{2}{c}{Trajectory Error $\downarrow$}
& Col.
& Off.
& Tra.\\
\cmidrule(lr){3-6}
\cmidrule(lr){7-8}
& & Realism & Kinematic & Map-based & Interactive
& ADE & minADE
&(\%) $\downarrow$  &  (\%) $\downarrow$ & (\%) $\downarrow$   \\
\midrule

SMART
& Top-K
& 0.7630
& 0.4595
& 0.8949
& 0.7948
& 3.9990
& 1.9630
& 5.50
& 12.50
& 3.90 \\

CAT-K
& Top-K
& \textbf{0.7647}
& \textbf{0.4588}
& \textbf{0.8950}
& \textbf{0.7993}
& 3.8810
& \textbf{1.9850}
& 4.70
& 12.50
& 3.50 \\

CAT-K
& Auto-labeler
& 0.7611
& 0.4508
& 0.8919
& 0.7963
& 3.8850
& 2.3730
& 4.20
& 12.60
& 3.50 \\

\rowcolor{lightblue}
CAT-K
& CRAFT
& 0.7189
& 0.3420
& 0.8437
& 0.7583
& 3.5490
& 3.3430
& \textbf{3.90}
& \textbf{11.58}
& 2.30 \\

\rowcolor{lightblue}
SMART
& CRAFT
& 0.7190
& 0.3493
& 0.8579
& 0.7630
& \textbf{3.4400}
& 3.1430
& 4.00
& 11.88
& \textbf{2.10} \\

\bottomrule
\end{tabular}
}
\label{tab:wosac_results}
\vspace{-1em}
\end{table}

\begin{table}[t]
\centering
\caption{
Simulation performance comparison under \textbf{behavioral rationality metrics}. Spd., Ang., and Dist. denote speed, angular speed, and nearest-agent distance; 
P-Ag. and P-Sc. denote per-agent and per-scene rates, respectively.
}
\resizebox{\linewidth}{!}{
\begin{tabular}{l l c c c c c c c c c}
\toprule
\multirow{2}{*}{Base Model}
& \multirow{2}{*}{Strategy}
& \multirow{2}{*}{Reference}
& \multicolumn{3}{c}{JSD ($\times 10^{-2}$) $\downarrow$}
& \multicolumn{2}{c}{Collision (\%) $\downarrow$}
& \multicolumn{2}{c}{Offroad (\%) $\downarrow$}
& Traffic (\%) $\downarrow$ \\
\cmidrule(lr){4-6}
\cmidrule(lr){7-8}
\cmidrule(lr){9-10}
\cmidrule(lr){11-11}
& & & Spd. & Ang. & Dist. & P-Ag. & P-Sc. & P-Ag. & P-Sc. & P-Sc. \\
\midrule

Log
& Log replay
& -- 
& 0.00 & 0.00 & 0.00
& 0.56 & 5.20
& 1.59 & 14.40
& 20.70 \\

\midrule
GUMP
& Top-K 
& ECCV2024
& 4.78 & 5.41 & 11.36
& 4.06 & 36.30
& 3.80 & 26.90
& 32.70 \\

SMART
& Top-K 
& NeurIPS2024
& 1.07 & 3.23 & 0.56
& 3.13 & 15.10
& 2.52 & 19.10
& 38.80 \\

CAT-K
& Top-K 
& CVPR2025
& 1.02 & 2.96 & \textbf{0.53}
& 3.41 & 15.70
& 2.54 & 19.50
& 36.50 \\

R1Sim
& Top-K 
& IEEE R-AL
& 1.05 & 3.16 & 0.56
& 3.36 & 15.50
& 2.23 & 16.30
& 36.60 \\

\midrule
CAT-K
& Auto-labeler 
& -
& 1.19 & 3.02 & 0.67
& 3.21 & 15.50
& 2.57 & 20.30
& 36.70 \\

\rowcolor{lightblue}
CAT-K
& CRAFT 
& -
& 0.92 & \textbf{2.20} & 0.69
& \textbf{2.46} & \textbf{10.80}
& 2.26 & \textbf{16.60}
& 26.60 \\

\rowcolor{lightblue}
SMART
& CRAFT 
& -
& \textbf{0.91} & \textbf{2.20} & 0.72
& 2.60 & 11.90
& \textbf{2.06} & 17.10
& \textbf{25.90} \\

\bottomrule
\end{tabular}
}
\vspace{-1em}
\label{tab:main_results}
\end{table}
\begin{figure}[t]
  \centering
    \includegraphics[width=\linewidth]{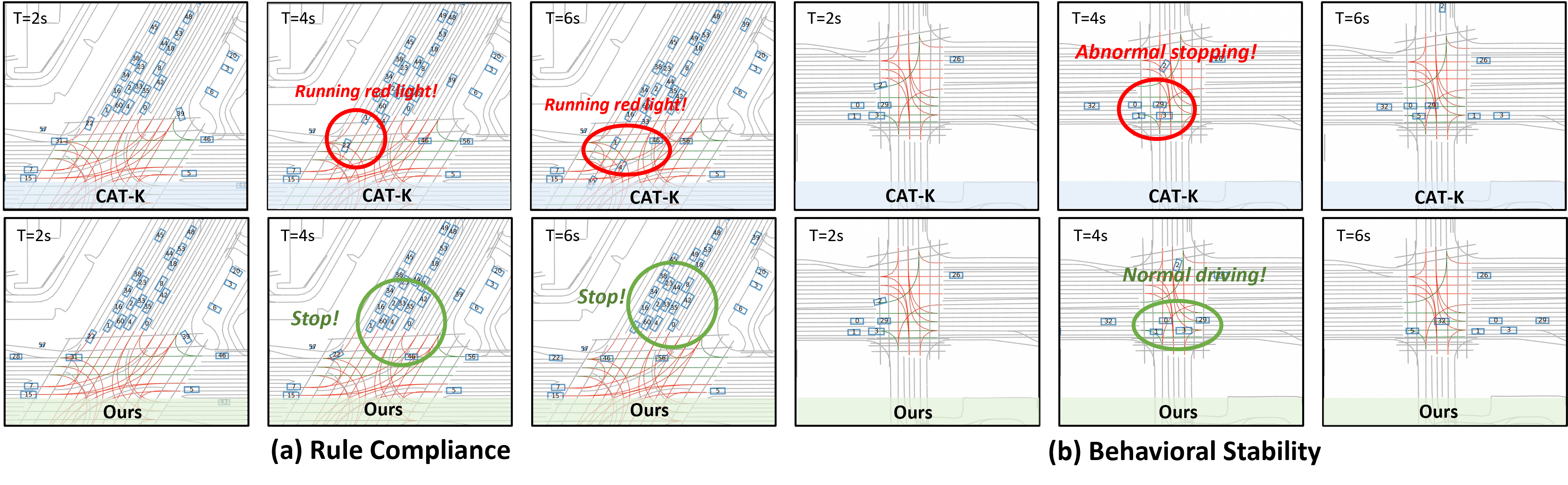}
  \caption{\textbf{Qualitative Comparison between CAT-K and CRAFT.} (a) CAT-K violates the traffic signal, while CRAFT reliably complies with the traffic rule. (b) CAT-K exhibits abnormal stopping and disrupts traffic flow, whereas CRAFT drives smoothly and maintains flow stability.
}
  \label{fig:visual}

\end{figure}
\subsection{Main result}

\textbf{WOSAC evaluation.}
Table~\ref{tab:wosac_results} reports the results under the WOSAC~\cite{montali2023waymo} protocol. 
CRAFT achieves fewer failures (i.e., collision, offroad, and traffic violation rates) at the all-agent level, while showing a certain degradation in the realism meta metric. 
We argue that this decrease should be interpreted carefully. 
WOSAC realism mainly reflects one-to-one similarity between the logged trajectories and simulated trajectories of scored agents. 
However, as discussed in Sec.~\ref{sec:intro}, locally observed logs may contain incomplete context-action associations due to perception boundaries and occlusions. 
As a result, this evaluation can favor behaviors that are close to the local log distribution, even when such behaviors may become abnormal under the complete simulated context. 
From this perspective, Top-K sampling~\cite{fan2018hierarchical} better preserves the base simulator's original ability to match logged trajectory patterns, leading to higher realism scores. 
In contrast, CRAFT guides decoding away from actions that are highly likely under the base model but potentially abnormal in closed-loop simulation. 
Therefore, its lower realism score reflects a trade-off between log-trajectory matching and global behavioral rationality, rather than a simple degradation in simulation quality. Further discussion on the WOSAC evaluation protocol and its limitations is provided in the Appendix \ref{sec:B}.

\textbf{Behavioral rationality evaluation.}
Furthermore, we evaluate behavioral rationality over all moving agents, including abnormal events ignored by WOSAC but critical for realistic traffic simulation. As shown in Table~\ref{tab:main_results}, CRAFT outperforms pretrained imitation-learning simulators such as GUMP \cite{hu2024solving} and SMART \cite{wu2024smart}, especially on collision and traffic-rule violation metrics. This indicates that pure log imitation captures local motion statistics but remains prone to abnormal behaviors. Post-training methods such as CAT-K \cite{zhang2025closed} and R1Sim \cite{wang2026learning} alleviate distribution shift through additional rollout- or reward-based training, yet remain tied to logged demonstrations. Consequently, their reductions in abnormal behavior rates remain limited. In contrast, the consistent improvements on both CAT-K and SMART demonstrate that CPE functions as a plug-in alignment module for different simulators, substantially improving the behavioral rationality of generated traffic agents. Directly using the auto-labeler's sparse rule-based outputs as online guidance provides limited candidate-level discrimination. In contrast, CPE amortizes these human-aligned criteria into dense contextual preference scores, enabling more stable probability calibration during decoding. CRAFT therefore improves closed-loop behavioral rationality while maintaining competitive distributional realism, with only a mild degradation in nearest-agent distance JSD, likely due to its preference for safer inter-agent spacing.

\textbf{Visualization.} A visual comparison between the CRAFT and the baseline method
is shown in Fig.\ref{fig:visual}. When the baseline fails to select a reasonable traffic behavior, CRAFT can suppress abnormal actions to maintain normal driving behavior.

\subsection{Analysis of Test-Time Alignment}
\begin{figure}[t]
\centering
\begin{minipage}[t]{0.42\linewidth}
  \centering
  \includegraphics[width=.9\linewidth]{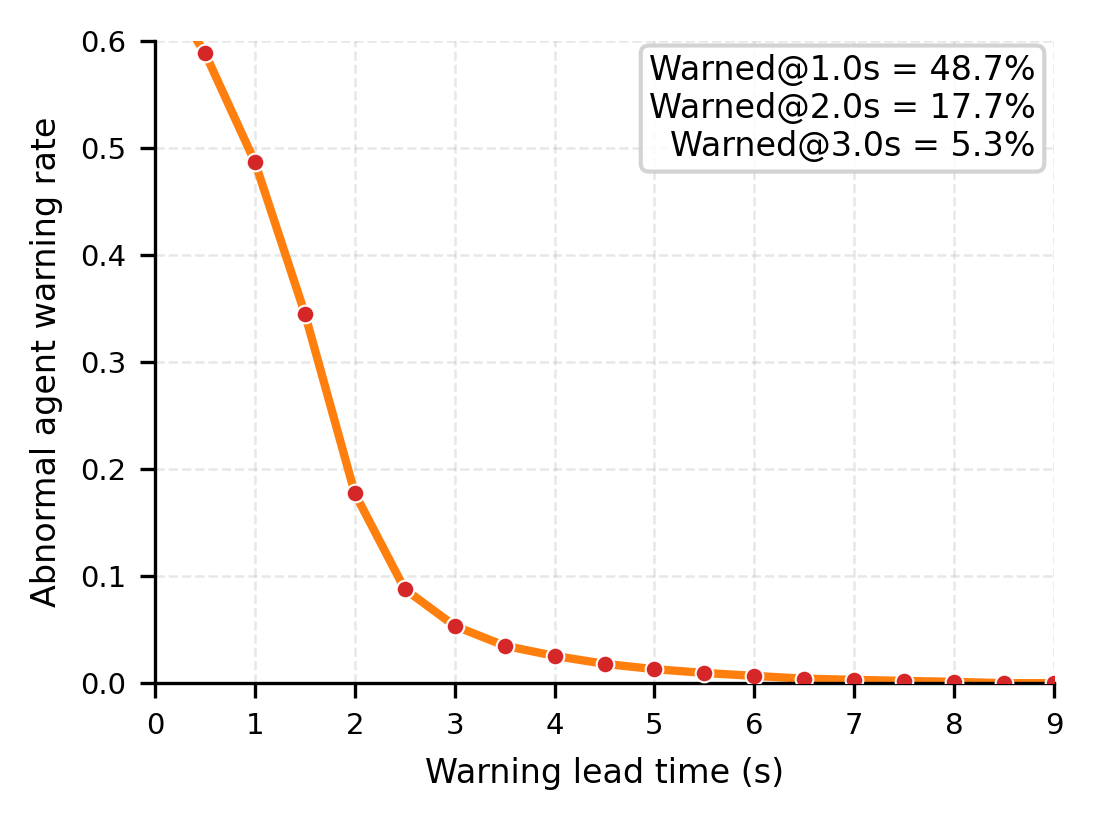}
  \caption{Abnormal agent warning rate of CPE over different warning lead times.}
  \label{fig:risk_warning}
\end{minipage}
\hfill
\begin{minipage}[t]{0.55\linewidth}
  \centering
  \includegraphics[width=.9\linewidth]{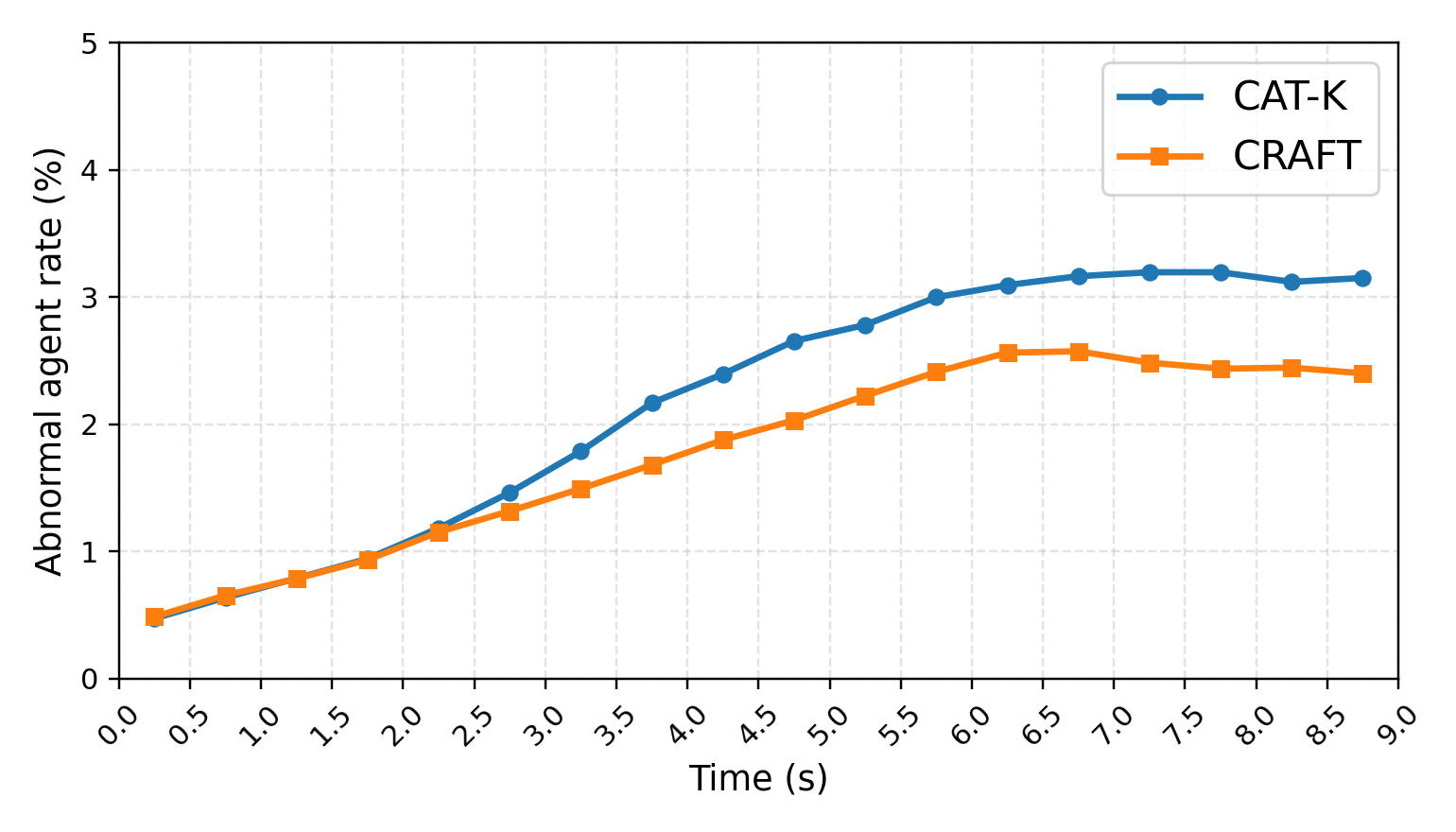}
  \caption{Abnormal agent rate curves of CRAFT against baseline over rollouts.}
  \label{fig:error_accumulation}
\end{minipage}
\vspace{-1em}
\end{figure}
\textbf{Proactive Risk Assessment.}
Fig.~\ref{fig:risk_warning} analyzes CPE's ability to anticipate abnormal behaviors. We define a warning as the first time step where the CPE score falls below $0.5$, and compare it with the actual occurrence time of abnormal actions. CPE warns $48.7\%$ of abnormal behaviors at least $1.0$s in advance and $17.7\%$ at least $2.0$s in advance, showing its ability to identify risky action distributions before failures materialize. This proactive signal enables test-time correction before execution, whereas rule-based evaluators usually react only after explicit violations have occurred.

\textbf{Error Accumulation Mitigation.}
As shown in Figure~\ref{fig:error_accumulation}, CAT-K suffers from evident error accumulation during closed-loop rollout, where abnormal behaviors become more frequent as autoregressive generation proceeds. After test-time calibration, CRAFT substantially reduces this risk, achieving a $23.79\%$ lower abnormal behavior rate at $9$s. This demonstrates that CPE-guided decoding can suppress risky candidates before execution and enhances the long-horizon behavioral rationality.

\begin{wrapfigure}{r}{0.50\textwidth}
  \captionof{table}{Runtime and behavioral performance of different test-time strategies.}
  \label{tab:runtime_analysis}
  \resizebox{\linewidth}{!}{%
\begin{tabular}{l c c c c}
\toprule
Strategy
& Col. (\%) $\downarrow$
& Off. (\%) $\downarrow$
& Tra. (\%) $\downarrow$
& Time (ms) $\downarrow$ \\
\midrule
Top-$K$
& 15.67
& 19.50
& 36.46
& \textbf{2.15} \\

Auto-labeler
& 15.50
& 20.30
& 36.70
& 491.77 \\

\rowcolor{lightblue}
CRAFT
& \textbf{10.80}
& \textbf{16.60}
& \textbf{26.60}
& 76.75 \\
\bottomrule
\end{tabular}
}
\end{wrapfigure}

\textbf{Runtime Overhead Analysis.} Table.~\ref{tab:runtime_analysis} compares different test-time alignment strategies in terms of behavioral performance and inference runtime. Top-K is computationally efficient but exhibits relatively high abnormal behavior rates. The rule-based auto-labeler provides only limited improvement while incurring a large runtime overhead. In contrast, CRAFT achieves consistently lower collision, offroad, and traffic violation rates with moderate additional cost.

\subsection{Ablation study}
Table~\ref{tab:ablation_results} analyzes the contribution of each component in CPE.

\textbf{Effect of Token-level Representation.}
Token-level representation is more suitable for CPE than sequence-level prediction. Since the base simulator performs autoregressive decoding over motion tokens, evaluating behaviors at the same granularity provides more direct supervision for test-time sampling. This leads to a stronger evaluator and more reliable closed-loop guidance.

\textbf{Effect of Complete-context What-if Data.}
Complete-context what-if rollouts are important for exposing simulator-induced failures. Logged data mainly provides static demonstrations, while many abnormal behaviors appear only after the simulator rolls out its own predictions. Training on what-if rollouts allows CPE to observe these failure modes under explicit scene context, improving its ability to guide the simulator away from unsafe or unreasonable actions.

\textbf{Effect of Preference Learning.}
The preference losses further improve CPE beyond BCE supervision. The intra-trajectory loss captures temporal degradation from normal behavior to failure, while the inter-rollout loss exploits alternative futures to learn relative behavioral preferences. Their combination provides complementary supervision and yields the most effective guidance, showing that CPE benefits from both absolute labels and group-level preference comparisons.

\begin{table}[t]
\centering
\caption{\textbf{Ablation study} of Contextual Preference Evaluator. Seq., Tok., and Rep. denote sequence-level, token-level, and representation, respectively.}
\resizebox{\linewidth}{!}{
\begin{tabular}{l l l c c c c c c c c c}
\toprule
\multicolumn{3}{c}{Components} 
& \multicolumn{4}{c}{Cls. Metrics} 
& \multicolumn{2}{c}{Collision (\%) $\downarrow$} 
& \multicolumn{2}{c}{Offroad (\%) $\downarrow$} 
& Traffic (\%) $\downarrow$ \\
\cmidrule(lr){1-3} 
\cmidrule(lr){4-7} 
\cmidrule(lr){8-9} 
\cmidrule(lr){10-11}
\cmidrule(lr){12-12}
Rep. & Data & Loss 
& Acc. & Rec. & Prec. & F1 
& P-Ag. & P-Sc. 
& P-Ag. & P-Sc. 
& P-Sc. \\
\midrule

Seq. &  $\mathcal{D}_{\mathrm{log}}$ & $\mathcal{L}_{bce}$ 
& 0.65 & 0.75 & 0.49 & 0.59 
& 4.36 & 20.50 & 3.10 & 25.80 & 42.30 \\

Tok. & $\mathcal{D}_{\mathrm{log}}$ & $\mathcal{L}_{bce}$ 
& 0.76 & 0.77 & 0.61 & 0.68 
& 4.23 & 19.90 & 3.01 & 23.60 &  38.70\\

Tok. & $\mathcal{D}_{\mathrm{pref}}$ & $\mathcal{L}_{bce}$ 
& 0.75 & 0.84 & 0.59 & 0.69 
& 3.10 & 15.00 & 2.37 & 17.30 & 36.10 \\

Tok. & $\mathcal{D}_{\mathrm{pref}}$  & $\mathcal{L}_{bce} + \mathcal{L}_{intra} $
& 0.78 & \textbf{0.85} & 0.65 & 0.74 
& 3.08 & 14.40 & 2.31 & 16.90 & 32.80 \\

Tok. & $\mathcal{D}_{\mathrm{pref}}$  & $\mathcal{L}_{bce} + \mathcal{L}_{inter}$ 
& \textbf{0.87} & 0.77 & 0.70 & 0.73 
& 2.93 & 12.80 & 2.62 & 19.40 & 28.20 \\

Tok. & $\mathcal{D}_{\mathrm{pref}}$  & $\mathcal{L}_{CPE}$ 
& 0.85 & 0.83 & \textbf{0.73} & \textbf{0.78} 
& \textbf{2.46 }& \textbf{10.80} & \textbf{2.26} & \textbf{16.60} & \textbf{26.60} \\

\bottomrule
\end{tabular}
}
\label{tab:ablation_results}

\end{table}

\begin{figure}
  \centering
  \includegraphics[width=.78\linewidth]{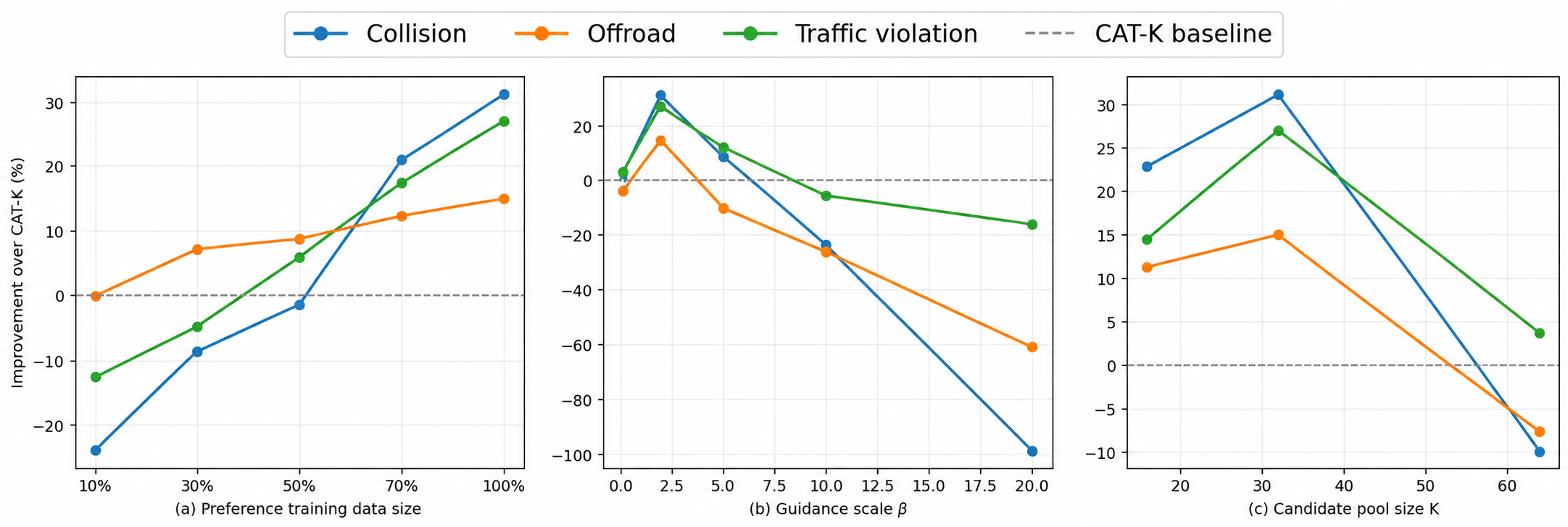}
    \caption{Results are reported as relative improvement over the CAT-K baseline under different preference training data sizes, guidance scales $\beta$, and candidate pool sizes $K$.
  }
    \label{fig:robustness}
    \vspace{-0.5em}
\end{figure}

\textbf{Robustness to Training and Inference Configurations.} Fig.~\ref{fig:robustness} shows that CRAFT is robust to both training and inference configurations. Increasing the preference training data consistently improves all behavioral metrics, indicating that CPE benefits from broader coverage of simulator-induced failure modes. For test-time alignment, a moderate guidance scale achieves the best trade-off: a small $\beta$ provides limited correction, while an overly large $\beta$ over-calibrates the base distribution and degrades performance. A similar trend is observed for the candidate size, where increasing $K$ from 16 to 32 improves guidance quality, but a larger pool introduces noisier candidates and hurts stability. 

\section{Conclusion}

In this paper, we present CRAFT, a test-time preference alignment framework centered on the Contextual Preference Evaluator (CPE). 
CRAFT learns rationality-aware preferences under complete scene context by using the frozen base simulator as a globally observable sandbox, where self-generated what-if rollouts expose failures induced by incomplete context learning. 
At inference time, CPE acts as a plug-in guidance module that recalibrates autoregressive decoding to suppress abnormal behaviors without retraining the simulator. 
Extensive experiments show that CRAFT improves closed-loop behavioral rationality while maintaining competitive simulation realism.

{
\bibliographystyle{neurips_2026}
\bibliography{neurips_2026}
}
\newpage
\appendix

In this supplementary material, we provide additional details in a question-driven manner to support the methodology and experiments discussed in the main manuscript:

\begin{itemize}
    \item \textbf{How is the complete experimental pipeline implemented?} 
    Appendix~\ref{sec:A} provides implementation details, including dataset configurations, trajectory discretization, model architectures, training setups, and inference procedures.

    \item \textbf{How do we validate behavioral alignment?} 
    Appendix~\ref{sec:B} details the validation protocols, annotation criteria, and evaluation settings used to assess whether generated behaviors are aligned with human-aligned driving priors.

    \item \textbf{What does the globally observable preference dataset contain, and what can we learn from it?} 
    Appendix~\ref{sec:C} presents the grouped preference dataset statistics, failure-type analysis, and representative visual examples, showing how simulator-induced failures are transformed into preference supervision.

    \item \textbf{What further insights can be obtained from additional qualitative and quantitative results?} 
    Appendix~\ref{sec:D} provides extended metric comparisons, alignment output visualizations, qualitative case studies, and failure case analysis.

    \item \textbf{What are the limitations and future directions?} 
    Appendix~\ref{sec:E} discusses the limitations of the current framework and outlines promising directions for future work.

    \item \textbf{Supplementary videos.} 
    Appendix~\ref{sec:F} describes the organization of the supplementary videos, which visualize the behavior alignment process and closed-loop simulation performance.
\end{itemize}

\section{How is the complete experimental pipeline implemented?}
\label{sec:A}
\subsection{Dataset and Splits}
Our experiments are conducted on the Waymo Open Motion Dataset (WOMD) \cite{ettinger2021large}, a widely used benchmark for motion prediction and traffic simulation, which includes 486995 training scenarios, 44097 validation scenarios and 44920 testing scenarios. Unlike simpler datasets that primarily focus on ego-vehicle routing, WOMD is uniquely characterized by its dense, highly interactive multi-agent environments. The dataset captures complex urban road topologies alongside dynamically interacting road users (e.g., vehicle, pedestrian, cyclist). This rich distribution of joint behaviors  (e.g., yielding, overtaking, and collision avoidance) makes it suitable for training and evaluating multi-agent traffic simulation models. In this dataset, each scenario provides 1 second of historical observations and an 8 seconds prediction horizon, recorded at a frequency of 10 Hz.

\subsection{Data preprocessing}
Following SMART~\cite{wu2024smart} and CAT-K~\cite{zhang2025closed}, we first construct discrete token vocabularies for both map elements and motion trajectories using k-disk clustering. Specifically, the map vocabulary contains 1,024 tokens, where each token represents a 5-meter polyline composed of 10 consecutive lane segments. The motion vocabulary contains 2,048 tokens, where each token corresponds to a 0.5-second trajectory fragment.

During data loading, we discretize the vectorized map elements in each scene into polylines of the same length and match them to the predefined map vocabulary according to the nearest-distance criterion. Motion trajectories are discretized in a similar manner by matching each trajectory fragment to its nearest motion token in the vocabulary. This tokenization procedure converts continuous map and motion information into discrete representations that can be directly used by the autoregressive simulator.

To ensure data quality and maintain a reasonable distribution of agent behaviors, we use all agents present at the 1-second timestamp as valid scene agents. For each scene, we select up to 32 agents within a 100-meter radius around the ego vehicle as training agents. If fewer than 32 agents are available within this range, all valid agents are retained; otherwise, 32 agents are randomly sampled from the candidate set. This filtering strategy preserves the most relevant local interactions around the ego vehicle while keeping the input size manageable for training.

\subsection{Model Architecture}
\begin{figure}
  \centering
  \includegraphics[width=.78\linewidth]{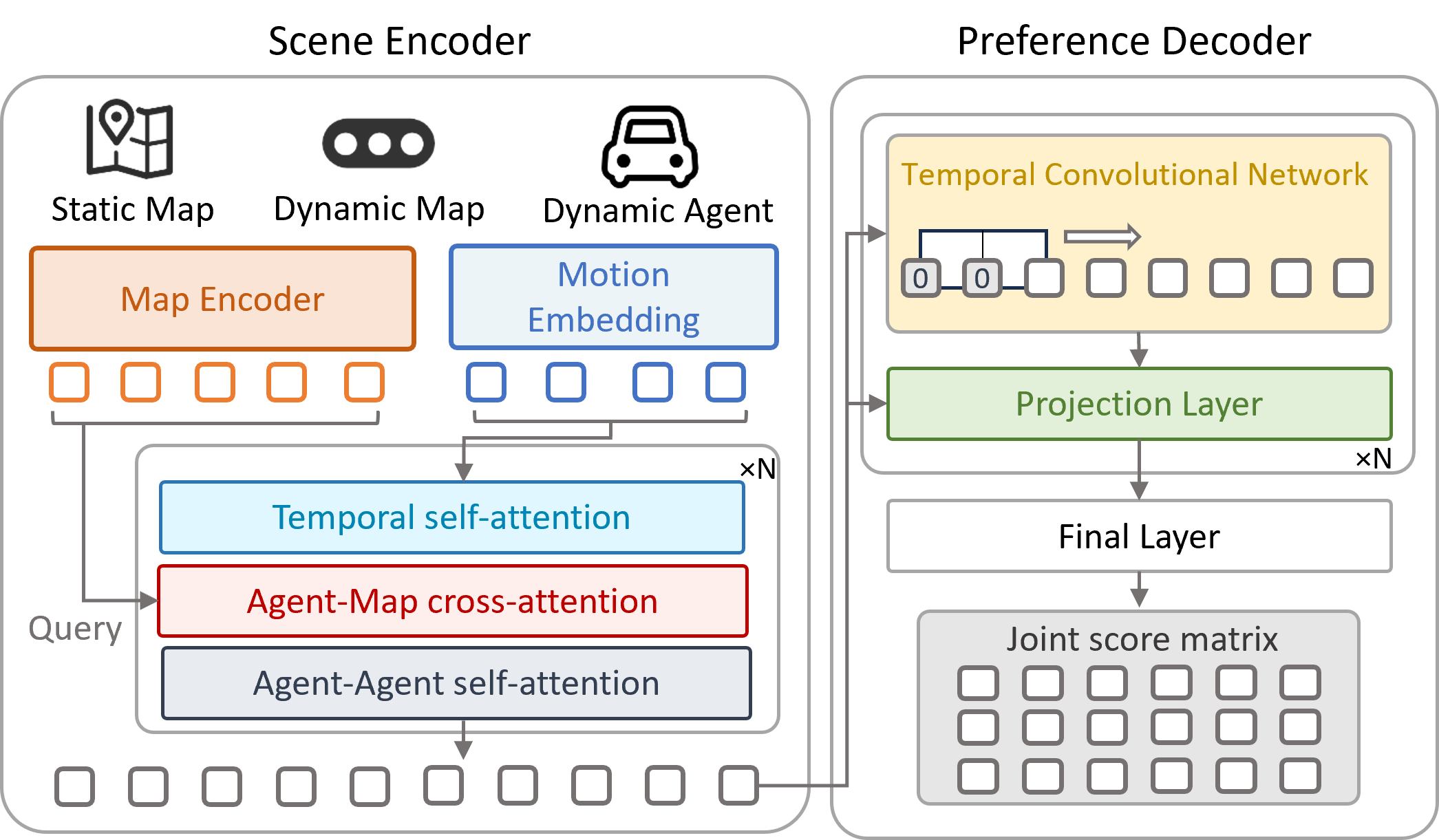}
    \caption{Contextual Preference Evaluator Architecture 
  }
    \label{fig:cpe_model}
    \vspace{-0.5em}
\end{figure}
Our model consists of two main components: a scene encoder and a lightweight decoder. 
The scene encoder follows a SMART-like architecture based on Transformer encoder blocks, which encode the tokenized map elements, agent histories, and scene-level interaction context into contextualized feature representations. 
This design allows the model to capture multi-agent dependencies and map-aware spatial relationships under the complete scene context.

On top of the encoded scene features, we employ a lightweight decoder implemented as a residual convolutional network. 
Specifically, the decoder is built from multiple temporal 1D convolutional layers with residual connections, enabling efficient modeling of local temporal patterns in candidate motion sequences. 
Compared with a heavy autoregressive Transformer decoder, this convolutional ResNet-style decoder provides a compact and efficient scoring module for contextual preference evaluation.

\subsection{Model Training}
\textbf{CPE Training.} The proposed method is implemented based on the open-source CAT-K codebase~\cite{zhang2025closed}. 
We retain the architectural configuration and training settings defined in the original repository to ensure consistency with the baseline implementation. 
The Contextual Preference Evaluator is initialized with the pretrained CAT-K model, which serves as the scene encoder and provides stable and discriminative scene representations. 
During training, the parameters of the scene encoder are kept frozen, and only the decoder of the Contextual Preference Evaluator is optimized. 
This design improves computational efficiency while reducing the overall training cost.

\textbf{Baseline Training.} We compare our method with several representative traffic simulation models, including SMART~\cite{wu2024smart}, GUMP~\cite{hu2024solving}, CAT-K \cite{zhang2025closed} and R1Sim~\cite{wang2026learning}. 
For these baselines, we use the official open-source implementations when available; otherwise, we implement the core algorithmic designs according to the original papers. To ensure a fair evaluation, all models are trained on the same training set.

\subsection{Hardware Setup}
To ensure a fair comparison, all methods are implemented and evaluated under the same experimental protocol. 
Both pretraining and fine-tuning are conducted on a cluster server with 4 NVIDIA RTX 4090 GPUs. 
Closed-loop simulation and metric evaluation are performed on a separate workstation equipped with 1 NVIDIA RTX 4090 GPU, an Intel Core i9-14900K CPU, and 128 GB RAM.

\subsection{Hyperparameters}
Table \ref{tab:hyperparams} summarizes the hyperparameters used in our approach, including the training settings of the Contextual Preference Evaluator (e.g., learning rate, batch size and number of epochs), the weighting coefficients of the loss functions, and the hyperparameters associated with the alignment strategy.
\begin{table*}[t]
\centering
\caption{Hyperparameter settings.}
\label{tab:hyperparams}
\renewcommand{\arraystretch}{1.08}
\setlength{\tabcolsep}{5pt}
\small
\begin{tabular}{l c | l c}
\toprule
\multicolumn{2}{c|}{\textbf{Model Architecture}} 
& \multicolumn{2}{c}{\textbf{Optimization and Loss}} \\
\midrule
Component & Value & Component & Value \\
\midrule
\multicolumn{2}{l|}{\textbf{General Architecture}} 
& \multicolumn{2}{l}{\textbf{Alignment Setting}} \\
Hidden dimension $d$ & 128 
& Weight of posterior probability $\beta$ & 2.0 \\
Number of attention heads & 8 
& Top-$k$ candidate size $k$ & 32 \\
Attention head dimension & 16 
& Top-$M$ selection size $M$ & 64 \\
Number of Fourier frequency bands & 64 
& Temp $t$ & 1.0 \\
Dropout rate & 0.1 
&  &  \\
\midrule
\multicolumn{2}{l|}{\textbf{Map Encoder}} 
& \multicolumn{2}{l}{\textbf{Training Setting}} \\
Map-map radius $r_{\mathrm{pl2pl}}$ & 10 
& Learning rate $\eta$ & $1\times10^{-5}$ \\
Max. map-map neighbors & 100 
& Batch size $B$ & 8 \\
Number of attention layers $L$ & 3 
& Number of epochs $E$ & 32 \\
\midrule
\multicolumn{2}{l|}{\textbf{Agent Encoder}} 
& \multicolumn{2}{l}{\textbf{Loss Function}} \\
Number of historical steps & 11 
& Ranking loss margin $m$ & 0.5 \\
Number of future steps & 80 
& Weight of BCE loss $\lambda_{1}$ & 0.5 \\
Number of attention layers $L$ & 6 
& Weight of intra-trajectory loss $\lambda_{2}$ & 1.5 \\
Temporal span & 30 
& Weight of inter-trajectory loss $\lambda_{3}$ & 1.0 \\
Temporal stride & 5 
&  &  \\
Map-agent radius $r_{\mathrm{pl2a}}$ & 30 
&  &  \\
Agent-agent radius $r_{\mathrm{a2a}}$ & 60 
&  &  \\
Max. map-agent neighbors & 300 
&  &  \\
Max. agent-agent neighbors & 300 
&  &  \\
Number of agent motion tokens & 2048 
&  &  \\
\midrule
\multicolumn{2}{l|}{\textbf{Decoder Architecture}} 
&  &  \\
Dropout rate & 0.15 
&  &  \\
Number of temporal conv. layers & 5 
&  &  \\
Temporal convolution kernel size & 6 
&  &  \\
GroupNorm groups & 16 
&  &  \\
Number of MLP layers & 4 
&  &  \\
Output dimension & 1 
&  &  \\
\bottomrule
\end{tabular}
\end{table*}

\section{How do we validate behavioral alignment?}
\label{sec:B}
\subsection{Evaluation protocol}
For evaluation, all methods are tested under a unified experimental setting. Given an initial 1-second scene observation, each model autoregressively predicts the trajectories of all agents for the following 8 seconds. The generated trajectories are then used to evaluate both the realism of the synthesized scenes and the behavioral rationality of the agents. During the framework validation stage, we randomly sample 2\% of the validation split to perform evaluation. Empirical observations \cite{rowe2025ctrl,zhang2025closed} show that this reduced validation subset maintains a distribution consistent with that of the original training and validation data, allowing it to serve as a reliable proxy for performance assessment while reducing computational cost. 

\subsection{Evaluation Metrics}
We adopt two complementary evaluation benchmarks to assess different aspects of closed-loop traffic simulation. 
The first follows the official WOSAC evaluator, which measures distributional realism and safety-related event rates by comparing simulated rollouts with logged futures. 
The second focuses on behavioral rationality, evaluating whether generated agents exhibit realistic behavior distributions and avoid abnormal closed-loop events under our validation protocol.

\paragraph{WOSAC evaluation metrics.}
The Waymo Open Sim Agents Challenge (WOSAC) evaluates whether a simulator can generate realistic multi-agent futures from logged initial conditions~\cite{montali2023waymo}. 
For each scenario, the simulator generates 32 stochastic rollouts, and the official evaluator compares these simulated futures with the logged trajectories of selected valid agents. 
These evaluated agents are determined by the official validity masks, which favor objects with sufficiently long valid future states to ensure reliable metric computation.

The core WOSAC realism metrics are \textbf{likelihood-based}. 
For each metric component, the evaluator constructs an empirical distribution from the 32 simulated rollouts and computes the approximate negative log-likelihood (NLL) of the logged future under this distribution:
\begin{equation}
    \mathrm{NLL} = - \frac{1}{|\mathcal{D}|} 
    \sum_{i=1}^{|\mathcal{D}|} 
    \log p_{\mathrm{sim}}(s_{\geq 1,i} \mid s_{<1,i}),
\end{equation}
where $\mathcal{D}$ denotes the set of evaluated scenarios, $s_{<1,i}$ represents the initial observation of scenario $i$, and $s_{\geq 1,i}$ represents its logged future. 
Since the exact likelihood of high-dimensional multi-agent futures is intractable, WOSAC approximates $p_{\mathrm{sim}}$ by normalizing histograms constructed from the submitted simulation samples.

The realism score is organized into three major groups. 
The \textit{kinematic} likelihood includes linear speed, linear acceleration, angular speed, and angular acceleration. 
The \textit{map-based} likelihood includes distance to road edge, offroad behavior, and traffic-light violation in the 2025 evaluator. 
The \textit{interactive} likelihood includes distance to the nearest object, collision, and time-to-collision. 
Together, these components measure whether simulated rollouts match real-world driving distributions in terms of motion dynamics, map compliance, and agent-agent interaction.

In addition to likelihood-based scores, the official evaluator also reports \textbf{rate-based} metrics for safety-critical events. 
In our experiments, we report collision rate, offroad rate, and traffic violation rate following the official evaluation outputs. 
These rates are computed over \textit{all valid agents} in the simulated scenes. 
For traffic violations, we follow the 2025 WOSAC evaluator and count red-light violations, i.e., cases where a vehicle passes through an intersection under a red traffic signal. 
We also report \textbf{trajectory errors}, including ADE averaged over the 32 generated rollouts and minADE computed from the rollout closest to the logged future.

\paragraph{Behavioral rationality metrics.}
Unlike WOSAC, which evaluates a selected subset of valid agents through the official challenge protocol, our behavioral rationality metrics assess behavior distributions and abnormal event rates over all moving agents in simulated scenes. 
We focus on moving agents rather than all existing agents because stationary vehicles may contain detection noise or pre-existing overlaps, which can introduce substantial noise into rate-based metrics. 
Thus, on top of the WOSAC validity filtering, we further exclude non-moving vehicles to obtain a cleaner evaluation of closed-loop behavioral rationality.

Our evaluation consists of two parts: distribution-based metrics and rate-based metrics. 
For \textbf{distribution-based metrics}, we use the Jensen--Shannon Distance (JSD) to quantify the discrepancy between feature distributions extracted from real and simulated rollouts. 
Instead of focusing on one-to-one trajectory matching against logged futures, we aggregate behavioral features over the entire simulated scene and compare their overall distributions with those from real data. 
We compute JSD over three behavior-related features: speed, angular speed, and nearest-agent distance. 
Lower JSD indicates that the simulated rollouts better match the real-world behavior distribution.

\textit{Jensen--Shannon Distance.}
For a given feature, let $p$ and $q$ denote the normalized histograms computed from real and simulated rollouts, respectively. 
The JSD is defined as:
\begin{equation}
\mathrm{JSD}(p, q) = 
\sqrt{
\frac{
D_{\mathrm{KL}}(p \,\|\, m) + D_{\mathrm{KL}}(q \,\|\, m)
}{2}
},
\end{equation}
where $m = \frac{1}{2}(p + q)$, and $D_{\mathrm{KL}}$ denotes the Kullback--Leibler divergence. 
Based on this definition, Spd JSD measures the discrepancy between linear speed distributions, Ang JSD measures the discrepancy between angular speed distributions, and Dist JSD measures the discrepancy between nearest-agent distance distributions.

For \textbf{rate-based metrics}, we report both scene-level and agent-level ratios, denoted as $P\text{-}\mathrm{Sc}$ and $P\text{-}\mathrm{Ag}$, respectively. 
$P\text{-}\mathrm{Sc}$ measures the percentage of scenes in which at least one abnormal event occurs, while $P\text{-}\mathrm{Ag}$ measures the percentage of moving agents involved in such events across all evaluated scenes. 
We compute these ratios for collision and offroad behaviors, and additionally report scene-level traffic violation rates. 
Together, these metrics provide a fine-grained assessment of behavioral rationality, complementing the likelihood-based realism evaluation of WOSAC.

\textit{Collision, offroad, and traffic-rule violation rates.}
Let $N_{\mathrm{sc}}$ denote the total number of evaluated scenes, and let $N_{\mathrm{ag}}$ denote the total number of evaluated moving agents. 
For a given event type, let $N_{\mathrm{event}}^{\mathrm{sc}}$ be the number of scenes in which the event occurs at least once, and let $N_{\mathrm{event}}^{\mathrm{ag}}$ be the number of moving agents that experience the event at least once. 
We define the scene-level ratio $P\text{-}\mathrm{Sc}$ and the agent-level ratio $P\text{-}\mathrm{Ag}$ as:
\begin{equation}
\begin{aligned}
P_{\mathrm{Sc}} &= \frac{N_{\mathrm{event}}^{\mathrm{sc}}}{N_{\mathrm{sc}}}, \\
P_{\mathrm{Ag}} &= \frac{N_{\mathrm{event}}^{\mathrm{ag}}}{N_{\mathrm{ag}}}.
\end{aligned}
\end{equation}

While $P\text{-}\mathrm{Sc}$ provides a macroscopic view of scenario-level safety, $P\text{-}\mathrm{Ag}$ offers a finer-grained view by measuring how many moving agents are involved in abnormal events. 
Following these definitions, we report $\mathrm{Collision}_{P\text{-}\mathrm{Sc}}$ and $\mathrm{Collision}_{P\text{-}\mathrm{Ag}}$, $\mathrm{Offroad}_{P\text{-}\mathrm{Sc}}$ and $\mathrm{Offroad}_{P\text{-}\mathrm{Ag}}$, as well as $\mathrm{Traffic}_{P\text{-}\mathrm{Sc}}$.

For collision detection, each moving agent is represented as an oriented bounding box, and collisions are identified using the Separating Axis Theorem. 

For offroad detection, we examine the four corners of the vehicle bounding box and determine whether any corner crosses the drivable boundary based on its vector orientation with respect to nearby road-edge segments. 
This box-level evaluation is stricter than center-point checking and captures partial vehicle departures from the drivable area.

For traffic violations, we build upon the official WOSAC evaluator and further refine the signal-compliance check by considering the permitted green-light direction. 
Specifically, a violation is counted not only when a vehicle runs a red light, but also when it remains stopped on an approach lane while the corresponding movement direction is given a green signal. 
This extension allows the metric to capture both unsafe red-light running and unreasonable green-light inaction, providing a more complete assessment of signal compliance in closed-loop simulation. 
In addition, We report traffic-rule violations only at the scene level. 
Unlike collision or offroad events, signal violations are localized to signalized intersections, and most agents in a scene are not subject to traffic-light constraints at a given time. 
Computing traffic violations at the agent level would therefore introduce a denominator bias and artificially dilute the metric. 
Moreover, traffic signals impose structural right-of-way constraints on the local traffic flow; a single red-light violation can invalidate the logical consistency of the intersection. 
Thus, scene-level traffic violation rate provides a more meaningful indicator of signal compliance.

\subsection{Additional Discussion on WOSAC Evaluation}

\begin{figure}
    \centering
    \includegraphics[width=\linewidth]{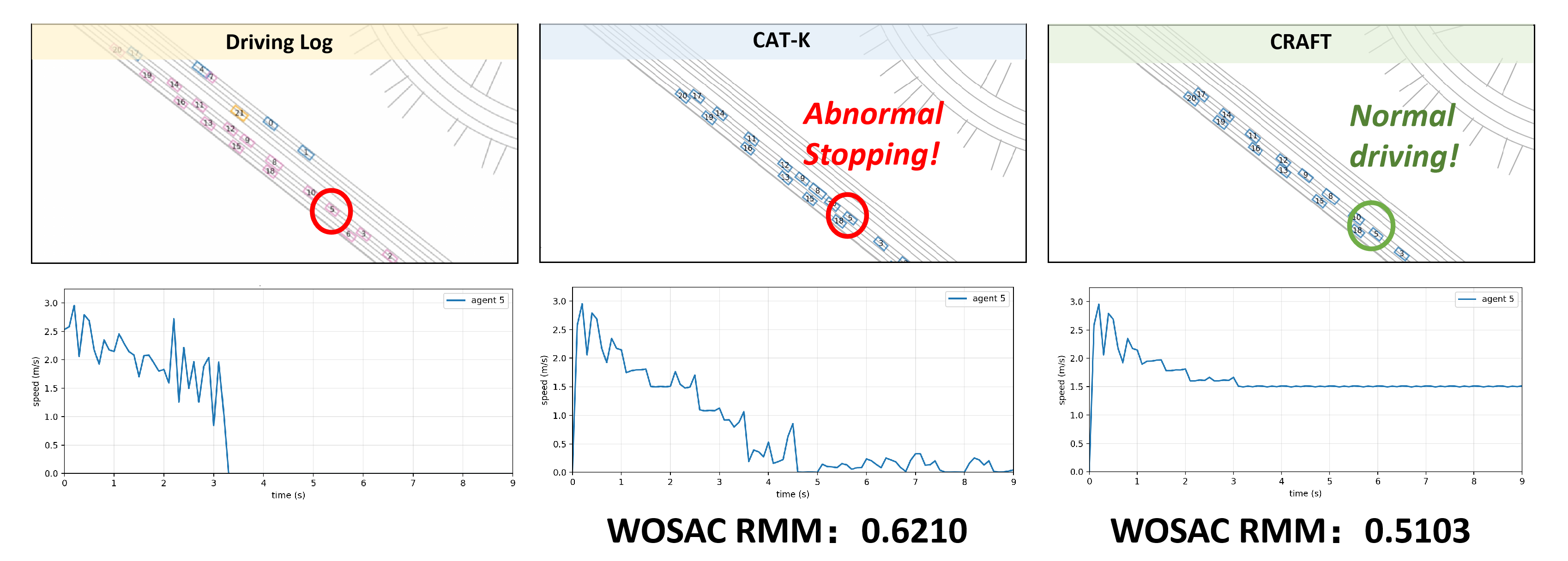}
\caption{
\textbf{Case study illustrating the limitation of WOSAC-style trajectory matching. }
In the driving log, the leading vehicle decelerates and disappears after approximately 4 seconds, leaving only a partial observation of its original intent. 
CAT-K closely imitates this pattern and eventually stops in the middle of the road, even though no leading vehicle exists in the simulated scene. 
In contrast, CRAFT uses the complete scene context to maintain forward motion, producing a more behaviorally rational rollout. 
Nevertheless, CAT-K receives a higher WOSAC realism score because it stays closer to the local logged trajectory pattern.
}
    \label{fig:wosac_case}
\end{figure}

To further illustrate the discussion in the main paper, we present a representative case study comparing the driving log, CAT-K, and CRAFT under the WOSAC protocol. 
This example helps explain why WOSAC may favor behaviors that remain closer to the local logged trajectory distribution, even when those behaviors become abnormal under the complete simulated context.

As shown in Fig.~\ref{fig:wosac_case}, the leading vehicle in the driving log first decelerates, likely because of a waiting vehicle or other downstream obstruction that is not fully visible in the recorded context. 
Moreover, this vehicle disappears from the log after approximately 4 seconds, so the logged trajectory only provides a partial observation of its original driving intent. 
CAT-K closely imitates this local behavior pattern and continues to decelerate in simulation. 
However, under the simulated scene, the corresponding vehicle no longer has a leading vehicle in front of it. 
As a result, CAT-K produces an unnatural behavior in which the vehicle gradually slows down and finally stops in the middle of the road.

In contrast, CRAFT takes the complete simulated scene into account during decoding. 
By recognizing that there is no vehicle ahead, CRAFT discourages the locally likely but globally inconsistent deceleration behavior and instead guides the vehicle to continue moving forward at a nearly constant speed. 
From the perspective of closed-loop behavioral rationality, this behavior is more consistent with common driving principles.

However, when we examine the WOSAC scores of these two simulated trajectories, CAT-K achieves a higher realism score at 0.6120, while CRAFT receives a lower one at 0.5103. 
This result reflects the central tension discussed in our paper: WOSAC rewards one-to-one similarity between simulated and logged trajectories of the scored agents, whereas CRAFT explicitly corrects behaviors that may be statistically consistent with local logs but irrational under global simulation. 
In this case, CAT-K is rewarded for closely mimicking the partial logged behavior, while CRAFT is penalized for maintaining a more globally reasonable driving policy. 
This example therefore highlights the mismatch between \emph{local log matching} and \emph{global closed-loop rationality}, and further motivates the use of complementary behavioral rationality metrics and qualitative analysis.

\section{What does the globally observable preference dataset contain, and what can we learn from it?}
\label{sec:C}
\subsection{Grouped What-if Scenario Construction}
As introduced in the main paper, human driving logs inherently suffer from a distributional coverage gap, primarily consisting of safe and compliant driving behaviors. To train the Contextual Preference Evaluator to recognize and penalize abnormal actions, we construct a large-scale what-if dataset, named \textbf{CRAFT15K}.

To construct the what-if scenario, we adopt a generation-then-evaluation paradigm. Specifically, we randomly sample 15,000 initial 1-second historical scenes from the WOMD dataset. Utilizing the base simulator (i.e., the pretrained CAT-K model), we conduct autoregressive simulations for each sampled scene. Driven by the multimodal characteristics of the traffic simulator and the temporal cascading effect inherent in the autoregressive generation process, applying diverse random seeds naturally yields a wide spectrum of what-if traffic scenarios. This streamlined approach ensures that the generated “what-if” trajectories authentically reflect the native capacity and behavioral distribution of the base model. By executing 32 independent random rollouts per scene, we ultimately construct the CRAFT15K dataset, providing a massive, diverse, and natively aligned substrate for subsequent dense step-level behavior data.

\begin{figure}
    \centering
    \includegraphics[width=0.9\linewidth]{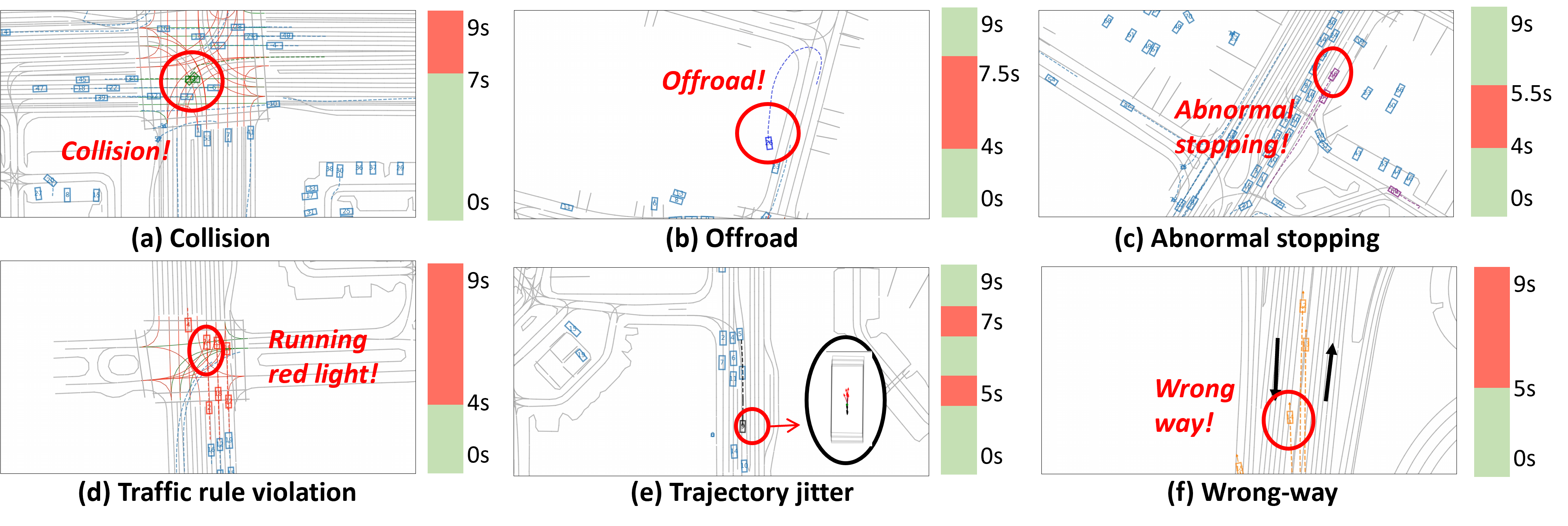}
\caption{\textbf{Visualization of partial negative-sample scenarios in CRAFT15K.}
The scenes are generated by our grouped complete-context what-if simulation pipeline. 
Vehicles with anomalous behaviors are color-coded as 
\textcolor{green}{green} (collision), 
\textcolor{darkblue}{dark blue} (offroad), 
\textcolor{purple}{purple} (abnormal stopping), 
\textcolor{red}{red} (traffic violation), 
\textcolor{black}{black} (front--rear jitter), and 
\textcolor{myyellow}{yellow} (wrong-way driving). 
\textcolor{blue}{Blue} indicates normal vehicles. 
The bar on the right represents the labels of the circled vehicles, 
where the \textcolor{red}{red} portion denotes negative labels and the \textcolor{green}{green} portion denotes positive labels.
}

    \label{fig:placeholder}
\end{figure}
\subsection{Annotation Protocol}
We construct behavior annotations for supervised learning by assigning step-wise labels to the generated grouped scene segments. Utilizing an automated evaluation pipeline, we assess the behavioral validity of each agent at every time step. To systematically evaluate these behaviors, we categorize six specific criteria into three macro-dimensions: \textit{Safety}, \textit{Compliance}, and \textit{Kinematics}. 

\noindent \textit{1) Safety}
\begin{itemize}
    \item \textbf{Collision} $(\mathcal{V}_{\mathrm{col}})$: We check whether an agent in the scene exhibits any explicit collision or hazardous spatial conflict with other agents over the entire time horizon. A time step is included in $\mathcal{V}_{\mathrm{col}}$ if no such violation is detected.
\end{itemize}

\noindent \textit{2) Compliance}
\begin{itemize}
    \item \textbf{Abnormal stopping} $(\mathcal{V}_{\mathrm{stop}})$: We check whether an agent undergoes substantial deceleration and comes to an evident stop from a normal driving state when there is no leading blocking vehicle, no red light, and no large-angle turning maneuver ahead. A time step is included in $\mathcal{V}_{\mathrm{stop}}$ if no such violation is detected.
    \item \textbf{Traffic rule violation} $(\mathcal{V}_{\mathrm{traffic}})$: We check whether an agent in a signal-controlled lane runs a red light or remains stationary when the signal is green. A time step is included in $\mathcal{V}_{\mathrm{traffic}}$ if no such violation is detected.
    \item \textbf{Offroad driving} $(\mathcal{V}_{\mathrm{offroad}})$: We check whether an agent drives beyond the boundary of the map. A time step is included in $\mathcal{V}_{\mathrm{offroad}}$ if no such violation is detected.
    \item \textbf{Wrong-way driving} $(\mathcal{V}_{\mathrm{wrong}})$: We check whether the direction of the lane traversed by an agent shows a large angular deviation from the agent's own heading direction. A time step is included in $\mathcal{V}_{\mathrm{wrong}}$ if no such violation is detected.
\end{itemize}

\noindent \textit{3) Kinematics} 
\begin{itemize}
    \item \textbf{Trajectory jitter} $(\mathcal{V}_{\mathrm{jitter}})$: We check whether an agent exhibits abnormal back-and-forth jittering in the lateral or longitudinal direction. A time step is included in $\mathcal{V}_{\mathrm{jitter}}$ if no such violation is detected.
\end{itemize}

Consequently, the final valid set $\mathcal{V}$ is strictly defined as the intersection of all individual criterion sets:
\begin{equation}
    \mathcal{V} = \mathcal{V}_{\mathrm{col}} \cap \mathcal{V}_{\mathrm{stop}} \cap \mathcal{V}_{\mathrm{traffic}} \cap \mathcal{V}_{\mathrm{offroad}} \cap \mathcal{V}_{\mathrm{wrong}} \cap \mathcal{V}_{\mathrm{jitter}}. 
\end{equation}

A trajectory segment is deemed valid (assigned a positive label) if and only if it strictly adheres to a predefined set of criteria; otherwise, it is labeled as negative. Accordingly, the binary label for agent $i$ at time step $t$ is formulated as:
\begin{equation}
    y_{i,t} = \begin{cases} 1, & (i,t) \in \mathcal{V},\\ 0, & (i,t) \notin \mathcal{V}. \end{cases} 
\end{equation}

\subsection{Annotation Visualization}
To facilitate understanding of the label definitions, we provide examples for each abnormal behavior in Fig.\ref{fig:placeholder}.

\begin{figure}
    \centering
    \includegraphics[width=\linewidth]{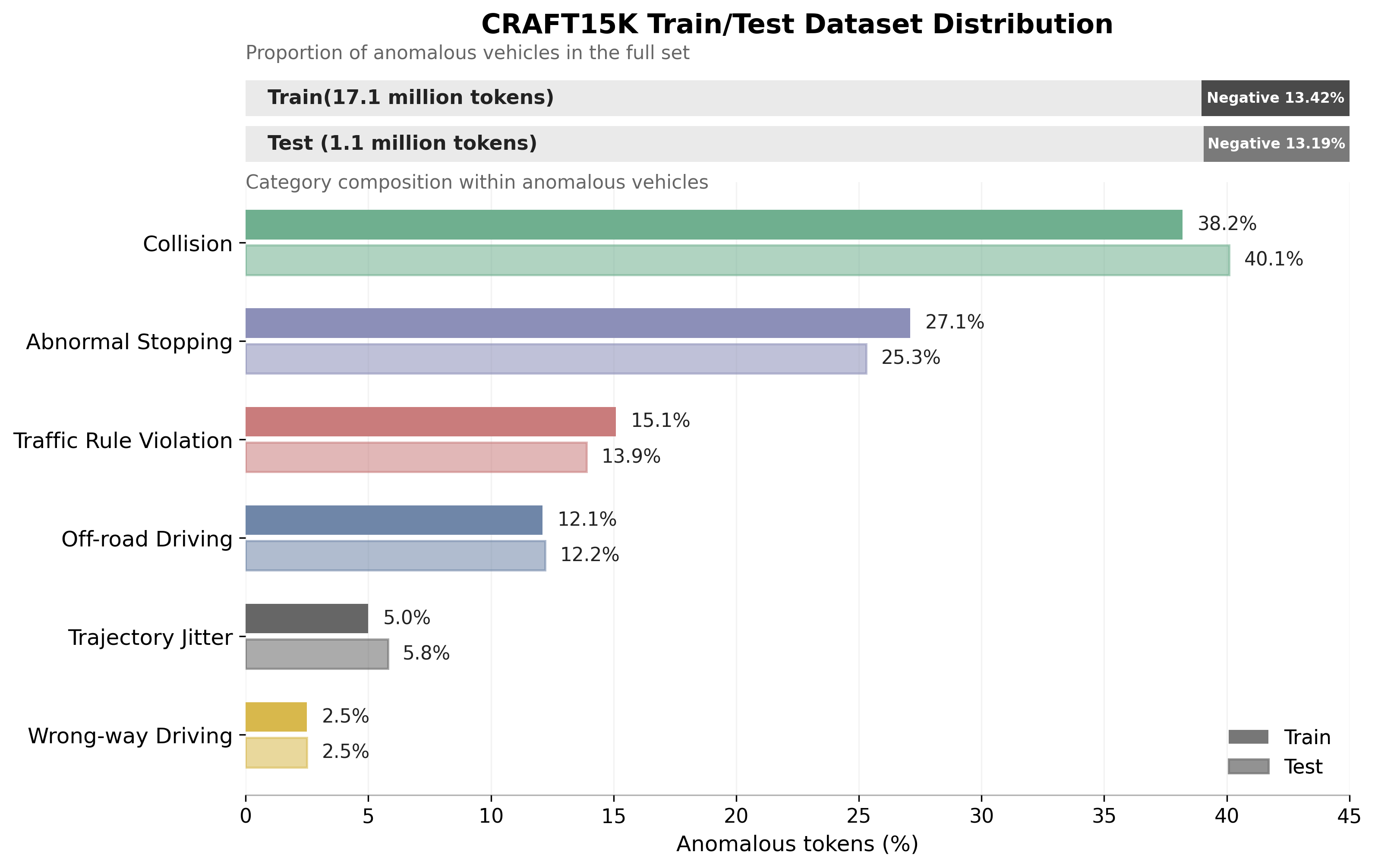}
    \caption{\textbf{Distribution of behavioral anomalies in the CRAFT15K dataset.} The chart visualizes the composition of anomalous tokens across a training set (17.1M tokens, 13.42\% negative) and a testing set (1.1M tokens, 13.19\% negative). The highly consistent distribution across both splits ensures reliable and unbiased evaluation for our predefined violation criteria.}

    \label{fig:dataset}
\end{figure}

\subsection{Annotation Statistics}

In the CRAFT15K dataset, our automated evaluation pipeline identifies anomalous behaviors across a substantial training set (17.1 million tokens) and a test set (1.1 million tokens). As shown in Fig.\ref{fig:dataset}, the overall proportion of negative samples remains highly consistent between the training (13.42\%) and testing (13.19\%) splits. Categorizing these anomalies reveals a distinct hierarchy that aligns with our evaluation dimensions, showing minimal distributional shift between the splits:
\begin{itemize}
    \item \textit{Safety:} Explicit collisions dominate the negative samples, accounting for a substantial 38.2\% in the training set and 40.1\% in the testing set.
    \item \textit{Compliance:} Violations in this category constitute a major portion, led by abnormal stopping (27.1\% train / 25.3\% test), followed by traffic rule violations (15.1\% train / 13.9\% test) and offroad driving (12.1\% train / 12.2\% test). Wrong-way driving remains a rare edge case, stable at 2.5\% across both sets.
    \item \textit{Kinematics: }Anomalies related to trajectory jitter account for the remaining 5.0\% of the training and 5.8\% of the testing negative steps.
\end{itemize}

Crucially, across the entire generated dataset, vehicles exhibiting at least one abnormal behavior consistently account for over 13\% of the total simulated scenarios. This dense concentration of boundary cases quantitatively validates our grouped rollout strategy. It demonstrates that our approach successfully bridges the distributional coverage gap inherent in driving logs, providing a sufficiently challenging, diverse, and high-quality substrate necessary for robust preference alignment.



\section{What further insights can be obtained from additional qualitative and quantitative results?}
\label{sec:D}

\subsection{Open-Loop Evaluation: Alignment Output Visualization}
\begin{figure}[h]
    \centering
    \includegraphics[width=0.9\linewidth]{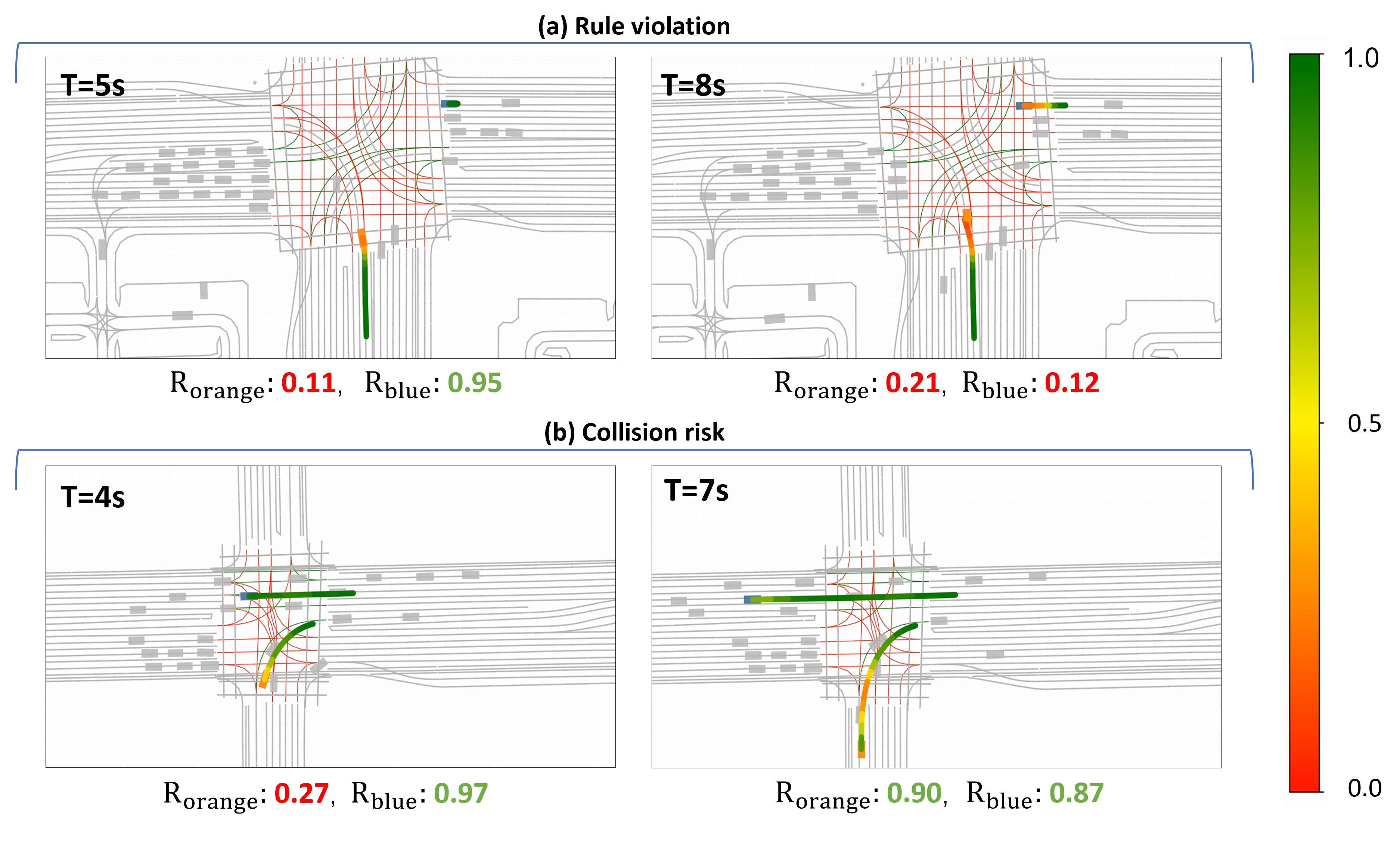}
\caption{\textbf{Qualitative evaluation of CAT-K generated trajectories using CRAFT.} (a) CRAFT accurately penalizes red-light infractions and unnatural lateral drifts. (b) The model consistently rewards stable car-following behaviors while assigning low scores to hazardous spatial conflicts.}
    \label{fig:score}
\end{figure}

This section presents visualizations of the step-wise evaluation scores assigned to agent behaviors by the Contextual Preference Evaluator across diverse driving scenarios in Fig.\ref{fig:score}. These qualitative results explicitly demonstrate the model's robust discriminative capacity in accurately identifying and penalizing anomalous or unsafe actions. Furthermore, they substantiate the underlying effectiveness of the proposed CRAFT evaluation method, proving its ability to provide reliable, interpretable assessments that correctly guide the agent decision process.

\textbf{Scenario 1: Rule violation.} 
This scenario highlights the model's sensitivity to rule grounding. At T=5s, the orange agent commits a severe red-light infraction, resulting in a heavily penalized score from CRAFT. Conversely, the blue agent exhibits strict signal compliance by yielding at the stop line, thereby receiving a high rationality score. By T=8s, the orange agent exhibits an unnatural lateral drift during its turning maneuver; our model immediately captures this kinematic anomaly, causing its score to drop even further. Meanwhile, the blue agent eventually violates the red light, which is also instantly penalized by a corresponding score reduction.

\textbf{Scenario 2: Collision risk.} 
This scenario highlights the model's sensitivity to spatial safety. At both T=4s and T=7s, the blue agent maintains a safe and stable headway, consistently earning high evaluation scores for its logical car-following dynamics. In contrast, the orange agent experiences a hazardous spatial conflict with surrounding vehicles at T=4s, leading to a sharply reduced score due to the imminent collision risk. As the scene evolves to T=7s, the orange agent safely disengages from the dense interaction with the gray vehicle and transitions back into a stable car-following state, which is correctly reflected by a swift recovery in its behavioral score.
\begin{figure}
    \centering
    \includegraphics[width=0.9\linewidth]{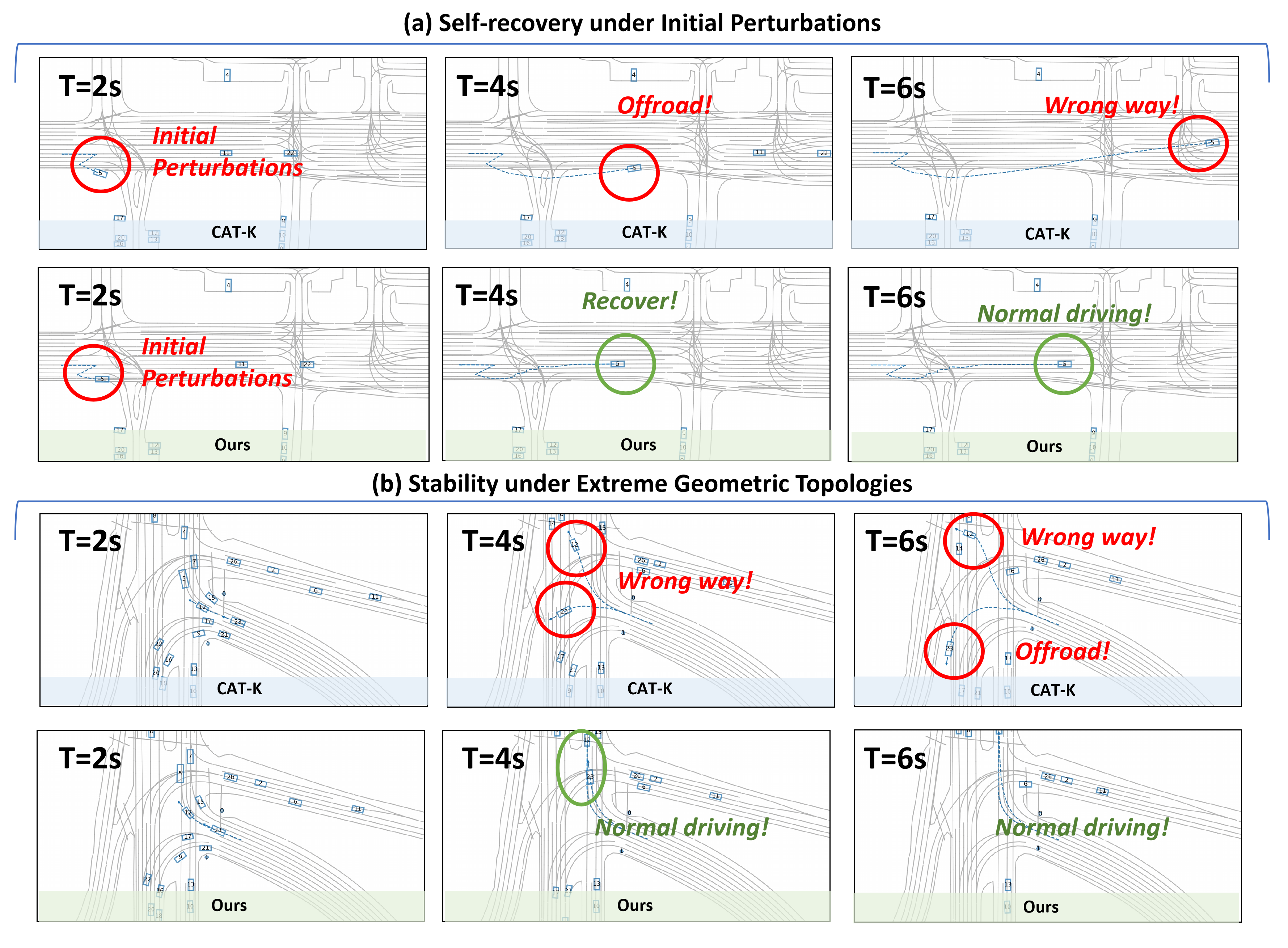}
    \caption{\textbf{Additional qualitative comparisons between CRAFT and CAT-K.} (a) Given an initial anomalous state, CRAFT gradually corrects the trajectory back to the lane center, whereas the baseline destabilizes into the oncoming lane. (b) CRAFT maintains strict lane adherence during a large-angle maneuver, successfully avoiding the offroad and wrong-way violations frequently triggered by the baseline's token selection errors.}
    \label{fig:additional_vis}
\end{figure}

\subsection{Closed-Loop Evaluation: Additional Qualitative Examples}

This section presents two additional complex scenarios generated by the CRAFT framework, which demonstrates its robustness against covariate shift and challenging road topologies during closed-loop autoregressive simulation in Fig.\ref{fig:additional_vis}.

\textbf{Scenario 1: Self-recovery under Initial Perturbations.} 
The first scenario evaluates the model's recovery capability when initialized with anomalous kinematic states. Due to the compounding nature of autoregressive generation, the baseline CAT-K model fails to solve out-of-distribution state, leading to causing it to generate abnormal behavior that ultimately forces the vehicle to drift unstably into the oncoming lane. Conversely, our Contextual Preference Evaluator exhibits strong self-recovery capabilities; it actively regularizes the generated tokens, safely guiding the vehicle to the lane center.

\textbf{Scenario 2: Stability under Extreme Geometric Topologies.} 
The second scenario highlights performance under strict spatial constraints, specifically during a large-angle turn. In the base simulator's motion token vocabulary, high-curvature motion tokens are naturally sparse. When confronted with such edge cases, the unaligned baseline frequently makes catastrophic token selection errors, resulting in offroad deviations or wrong-way driving. By integrating preference alignment, CRAFT significantly enhances the stability of these sharp maneuvers, ensuring that the selected tokens  keep the vehicle safely within its bounds.

\subsection{Offline CPE Ranking Analysis}

In addition to the online test-time guidance used in the main paper, we conduct an offline ranking analysis to better understand the roles of the base simulator and CPE. 
This experiment is not used as our main inference strategy; instead, it serves as a diagnostic study to examine whether the base simulator already contains potentially reasonable behaviors in its sampled trajectory set, and whether CPE can identify them under complete scene context.

Specifically, for each scenario, we first let the base simulator generate 64 closed-loop rollout candidates from the same logged initial state without applying CPE or auto-labeler guidance during decoding. 
After the rollouts are completed, CPE or auto-labeler evaluates each candidate trajectory under the full simulated scene context and assigns an overall preference score. 
We then select the top 32 scoring rollout as the final output. 
This differs from our online CRAFT inference, where CPE is applied at each autoregressive decoding step to reweight candidate actions before the trajectory is fully generated. 
Therefore, offline ranking only selects among already generated futures, while online guidance actively changes the generation process.

\begin{table}[t]
\centering
\caption{Simulation performance comparison under\textbf{ WOSAC metrics}. Col., Off., and Tra. denote collision rate, offroad rate, and traffic violation rate, respectively.}
\label{tab:sim_metrics}
\resizebox{\linewidth}{!}{
\begin{tabular}{l l c c c c c c c c c}
\toprule
\multirow{2}{*}{Base Model}
& \multirow{2}{*}{Strategy}
& \multicolumn{4}{c}{WOSAC Metrics $\uparrow$}
& \multicolumn{2}{c}{Trajectory Error $\downarrow$}
& Col.
& Off.
& Tra.\\
\cmidrule(lr){3-6}
\cmidrule(lr){7-8}
& & Realism & Kinematic & Map-based & Interactive
& ADE & minADE
&(\%) $\downarrow$  &  (\%) $\downarrow$ & (\%) $\downarrow$   \\
\midrule

SMART
& Top-K
& 0.7630
& 0.4595
& 0.8949
& 0.7948
& 3.9990
& 1.9630
& 5.50
& 12.50
& 3.90 \\

CAT-K
& Top-K
& 0.7647
& \textbf{0.4588}
& 0.8950
& 0.7993
& 3.8810
& 1.9850
& 4.70
& 12.50
& 3.50 \\

\rowcolor{lightred}
CAT-K
& Auto-labeler-rank
& \textbf{0.7663}
& \textbf{0.4588}
& \textbf{0.8960}
& \textbf{0.8017}
& 3.8760
& \textbf{1.9520}
& 4.30
& 12.50
& 3.30 \\

CAT-K
& Auto-labeler
& 0.7611
& 0.4508
& 0.8919
& 0.7963
& 3.8850
& 2.3730
& 4.20
& 12.60
& 3.50 \\
\rowcolor{lightred}
CAT-K
& CPE-rank
& 0.7654
& 0.4615
& 0.8947
& 0.7997
& 4.0020
& 1.9790
& 4.80
& 12.40
& 3.80 \\

\rowcolor{lightblue}
CAT-K
& CRAFT
& 0.7189
& 0.3420
& 0.8437
& 0.7583
& 3.5490
& 3.3430
& \textbf{3.90}
& \textbf{11.58}
& 2.30 \\
\rowcolor{lightred}
SMART
& CPE-rank
& 0.7643
& 0.4573
& 0.8949
& 0.7990
& 3.8830
& 1.9820
& 4.70
& 12.70
& 3.40 \\

\rowcolor{lightblue}
SMART
& CRAFT
& 0.7190
& 0.3493
& 0.8579
& 0.7630
& \textbf{3.4400}
& 3.1430
& 4.00
& 11.88
& \textbf{2.10} \\

\bottomrule
\end{tabular}
}
\vspace{-1em}
\end{table}

\begin{table}
\centering
\caption{
Simulation performance comparison under \textbf{behavioral rationality metrics}. Spd., Ang., and Dist. denote speed, angular speed, and nearest-agent distance; 
P-Ag. and P-Sc. denote per-agent and per-scene rates, respectively.
}
\resizebox{\linewidth}{!}{
\begin{tabular}{l l c c c c c c c c c}
\toprule
\multirow{2}{*}{Base Model}
& \multirow{2}{*}{Strategy}
& \multirow{2}{*}{Reference}
& \multicolumn{3}{c}{JSD ($\times 10^{-2}$) $\downarrow$}
& \multicolumn{2}{c}{Collision (\%) $\downarrow$}
& \multicolumn{2}{c}{Offroad (\%) $\downarrow$}
& Traffic (\%) $\downarrow$ \\
\cmidrule(lr){4-6}
\cmidrule(lr){7-8}
\cmidrule(lr){9-10}
\cmidrule(lr){11-11}
& & & Spd. & Ang. & Dist. & P-Ag. & P-Sc. & P-Ag. & P-Sc. & P-Sc. \\
\midrule

Log
& Log replay
& -- 
& 0.00 & 0.00 & 0.00
& 0.56 & 5.20
& 1.59 & 14.40
& 20.70 \\

\midrule
GUMP
& Top-K 
& ECCV2024
& 4.78 & 5.41 & 11.36
& 4.06 & 36.30
& 3.80 & 26.90
& 32.70 \\

SMART
& Top-K 
& NeurIPS2024
& 1.07 & 3.23 & 0.56
& 3.13 & 15.10
& 2.52 & 19.10
& 38.80 \\

CAT-K
& Top-K 
& CVPR2025
& 1.02 & 2.96 & \textbf{0.53}
& 3.41 & 15.70
& 2.54 & 19.50
& 36.50 \\

R1Sim
& Top-K 
& IEEE R-AL
& 1.05 & 3.16 & 0.56
& 3.36 & 15.50
& 2.23 & 16.30
& 36.60 \\

\midrule
\rowcolor{lightred}
CAT-K
& Auto-labeler-rank 
& -
& 1.09 & 3.15 & 0.55
& 2.72 & 12.80
& 2.39 & 17.90
& 34.30 \\

CAT-K
& Auto-labeler 
& -
& 1.19 & 3.02 & 0.67
& 3.21 & 15.50
& 2.57 & 20.30
& 36.70 \\

\rowcolor{lightred}
CAT-K
& CPE-rank 
& -
& 1.09 & 3.18 & 0.55
& 3.12 & 13.90
& 2.45 & 17.90
& 34.00 \\

\rowcolor{lightblue}
CAT-K
& CRAFT 
& -
& 0.92 & \textbf{2.20} & 0.69
& \textbf{2.46} & \textbf{10.80}
& 2.26 & \textbf{16.60}
& 26.60 \\

\rowcolor{lightred}
SMART
& CPE-rank 
& -
& 1.09 & 3.16 & 0.55
& 2.83 & 13.00
& 2.29 & 17.70
& 35.30 \\

\rowcolor{lightblue}
SMART
& CRAFT 
& -
& \textbf{0.91} & \textbf{2.20} & 0.72
& 2.60 & 11.90
& \textbf{2.06} & 17.10
& \textbf{25.90} \\

\bottomrule
\end{tabular}
}
\label{tab:main_results_add}
\end{table}

The offline ranking results provide two useful observations. 
First, CPE can consistently rank safer and more behaviorally rational rollouts higher, indicating that the learned preference scores capture meaningful global interaction quality rather than merely fitting the annotation labels. 
Second, the fact that offline ranking can already improve the results suggests that the base simulator has not fully exhausted its behavioral potential: reasonable futures may exist in its candidate set, but are not reliably selected by the original sampling strategy. 
This observation further motivates our test-time guidance design, which incorporates CPE earlier in the decoding process to steer the simulator toward such globally coherent behaviors before abnormal actions accumulate.

\subsection{Complete Sensitivity Experimental Results}
The main paper reports the overall comparison across different hyperparameter settings and baseline models. 
Here, we provide the complete numerical results in table.\ref{tab:ablation_all} to facilitate a more detailed examination of these analyses.

\begin{table}[t]
\centering
\caption{Ablation studies on CPE training size, guidance strength $\beta$, and candidate set size $K$.}
\label{tab:ablation_all}

\begin{subtable}[t]{0.32\linewidth}
\centering
\caption{Training size}
\label{tab:ablation_train_size}
\resizebox{\linewidth}{!}{
\begin{tabular}{lccccc}
\toprule
\multirow{2}{*}{Size}
& \multicolumn{2}{c}{Collision $\downarrow$}
& \multicolumn{2}{c}{Offroad $\downarrow$}
& Traffic $\downarrow$ \\
\cmidrule(lr){2-3}
\cmidrule(lr){4-5}
\cmidrule(lr){6-6}
& P-Sc & P-Ag & P-Sc & P-Ag & P-Sc \\
\midrule
10\%  & 0.1940 & 0.0410 & 0.1950 & 0.0252 & 0.4100 \\
30\%  & 0.1700 & 0.0359 & 0.1810 & 0.0241 & 0.3820 \\
50\%  & 0.1590 & 0.0287 & 0.1780 & 0.0239 & 0.3430 \\
70\%  & 0.1240 & 0.0253 & 0.1710 & 0.0231 & 0.3010 \\
100\% & \textbf{0.1080} & \textbf{0.0246} & \textbf{0.1660} & \textbf{0.0226} & \textbf{0.2660} \\
\bottomrule
\end{tabular}
}
\end{subtable}
\hfill
\begin{subtable}[t]{0.32\linewidth}
\centering
\caption{Guidance strength}
\label{tab:ablation_beta}
\resizebox{\linewidth}{!}{
\begin{tabular}{lccccc}
\toprule
\multirow{2}{*}{$\beta$}
& \multicolumn{2}{c}{Collision $\downarrow$}
& \multicolumn{2}{c}{Offroad $\downarrow$}
& Traffic $\downarrow$ \\
\cmidrule(lr){2-3}
\cmidrule(lr){4-5}
\cmidrule(lr){6-6}
& P-Sc & P-Ag & P-Sc & P-Ag & P-Sc \\
\midrule
0.2 & 0.1530 & 0.0303 & 0.2020 & 0.0258 & 0.3530 \\
2   & \textbf{0.1080} & \textbf{0.0246} & \textbf{0.1660} & \textbf{0.0226} & \textbf{0.2660} \\
5   & 0.1430 & 0.0323 & 0.2150 & 0.0288 & 0.3210 \\
10  & 0.1940 & 0.0439 & 0.2460 & 0.0332 & 0.3860 \\
20  & 0.3110 & 0.0669 & 0.3140 & 0.0366 & 0.4240 \\
\bottomrule
\end{tabular}
}
\end{subtable}
\hfill
\begin{subtable}[t]{0.32\linewidth}
\centering
\caption{Candidate set size}
\label{tab:ablation_topk}
\resizebox{\linewidth}{!}{
\begin{tabular}{lcccccc}
\toprule
\multirow{2}{*}{Strategy}
& \multirow{2}{*}{$K$}
& \multicolumn{2}{c}{Collision $\downarrow$}
& \multicolumn{2}{c}{Offroad $\downarrow$}
& Traffic $\downarrow$ \\
\cmidrule(lr){3-4}
\cmidrule(lr){5-6}
\cmidrule(lr){7-7}
& & P-Sc & P-Ag & P-Sc & P-Ag & P-Sc \\
\midrule
CAT-K & 16 & 0.1390 & 0.0298 & 0.1810 & 0.0254 & 0.3610 \\
CAT-K & 32 & 0.1567 & 0.0341 & 0.1950 & 0.0254 & 0.3646 \\
CAT-K & 64 & 0.1940 & 0.0424 & 0.1910 & 0.0248 & 0.3680 \\
\midrule
CRAFT & 16 & 0.1210 & 0.0267 & 0.1730 & 0.0238 & 0.3120 \\
CRAFT & 32 & \textbf{0.1080} & \textbf{0.0246} & \textbf{0.1660} & \textbf{0.0226} & \textbf{0.2660} \\
CRAFT & 64 & 0.1720 & 0.0378 & 0.2100 & 0.0273 & 0.3510 \\
\bottomrule
\end{tabular}

}
\end{subtable}
\end{table}

\section{What are the limitations and future directions?}
\label{sec:E}
\subsection{Failure Case}
\begin{figure}[h]
    \centering
    \includegraphics[width=\linewidth]{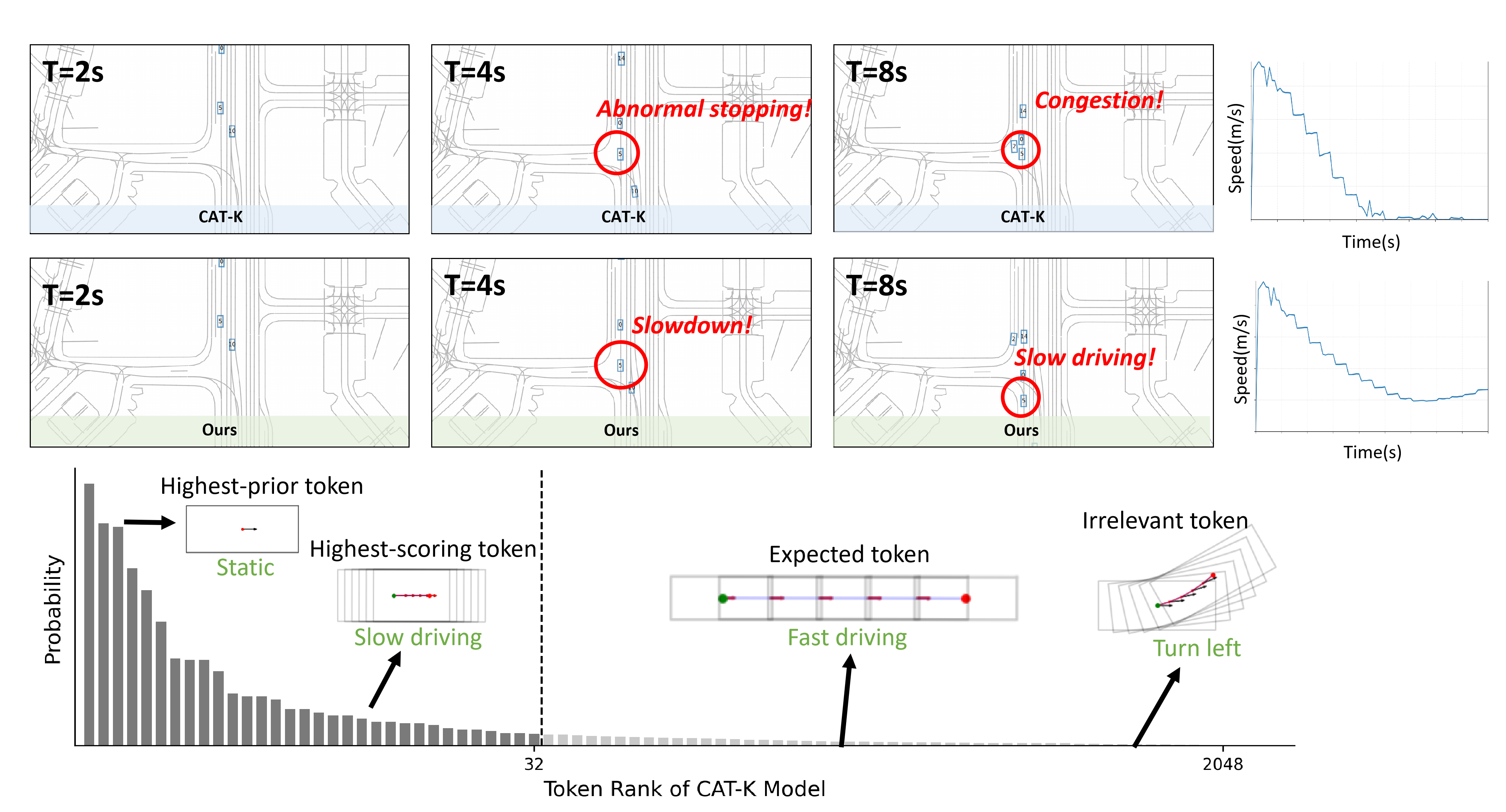}
    \caption{\textbf{Failure case due to candidate set limitations.} \textit{(Left)} Qualitative visualization at a T-junction, where both CAT-K and CRAFT exhibit anomalous behaviors. \textit{(Right)} The velocity-time profile of the red circle vehicle showing an unnatural deceleration. \textit{(Bottom)} The token probability distribution from the base simulator at T=4s, indicating that the optimal turning tokens are absent from the Top-32 candidate pool.}
    \label{fig:failure}
\end{figure}

While our framework improves scenario realism in most cases, we present a failure case to transparently discuss the limitations of our current sampling strategy. As illustrated in Fig.~\ref{fig:failure}, when crossing a T-junction, the vehicle generated by the baseline CAT-K exhibits an unnatural stop. Our CRAFT model shows a similar behavior, characterized by an unexpected deceleration instead of keeping moving.

Our analysis at step T=4s reveals that this failure stems from our reliance on the Top-k candidate set (k=32 in our implementation), which is adopted to maintain computational efficiency during the lookahead search. In this scene, the base simulator assigns disproportionately low probabilities to the optimal moving tokens, causing them to fall completely outside the Top-32 proposal pool. Consequently, the optimal actions are excluded from the alignment evaluation. The Contextual Preference Evaluator is forced to score and select from a set of suboptimal proposals, resulting in the safest available but unnatural deceleration. This case factually demonstrates that the performance of CRAFT is fundamentally bottlenecked by the quality of the provided candidate space. 

\subsection{Limitations and Future Work}

Although CRAFT demonstrates the effectiveness of test-time behavioral alignment for closed-loop traffic simulation, we acknowledge that its current inference strategy still leaves substantial room for improvement. 
In this work, the search and reweighting procedure is designed as a simple and experimental realization of preference-guided decoding. 
While effective in reducing abnormal behaviors, it may still be constrained by the quality and diversity of the candidate actions proposed by the base simulator. 
If essential motion tokens are not included in the candidate set, CPE can only re-rank the available choices rather than recover the missing behaviors.

Nevertheless, we believe this limitation also highlights an important direction for future traffic simulation research. 
Most existing methods focus on improving imitation learning or post-training the simulator, whereas our work suggests that correcting behavioral biases at test time can provide a lightweight and complementary path toward more reliable closed-loop simulation. 
This perspective is particularly relevant for practical simulators and world models, where retraining large generative models can be costly, unstable, or difficult to deploy.

Future work may explore more advanced decoding and search strategies, such as dynamic expansion of the candidate set, adaptive guidance strength or hybrid methods that combine preference-guided decoding with improvements to the base simulator's prior distribution. 
In particular, directly improving the base simulator's proposal distribution could help prevent essential tokens from being excluded before CPE evaluation, enabling stronger alignment between local driving priors and globally coherent behaviors.

\section{Supplementary videos}
\label{sec:F}
We provide a collection of supplementary videos to accompany this manuscript. These videos dynamically animate the static visualizations presented in both the main text and this appendix, offering additional qualitative evidence to substantiate the effectiveness of our proposed framework. The video files are organized into three distinct directories:

\begin{itemize}
    \item \textbf{\texttt{video/dataset/}}: This folder contains animated visualizations of the annotated samples from our CRAFT15K dataset. These videos explicitly highlight the various categories of anomalous agent behaviors (e.g., collisions, abnormal stopping, offroad driving) captured by our automated evaluation pipeline.
    
    \item \textbf{\texttt{video/Open\_loop\_evaluation/}}: This directory contains open-loop evaluation visualizations of our proposed CRAFT-r applied to the base simulator's trajectories. These dynamic demonstrations illustrate the Contextual Preference Evaluator's step-wise scoring of agent actions, clearly validating its precise discriminative capacity to accurately identify and penalize anomalous behaviors.

    \item \textbf{\texttt{video/Closed\_loop\_evaluation/}}: This folder features  closed-loop qualitative comparisons between our proposed CRAFT framework and the baseline models. Corresponding to the static figures presented in the manuscript, these rollouts clearly demonstrate the performance improvements achieved by our method in enhancing both behavioral rationality and realism across diverse traffic situations.

    \item \textbf{\texttt{video/failure/}}: Here, we provide videos of our failure cases. These videos illustrate the extreme boundary conditions and highly constrained scenarios that our current model cannot yet perfectly resolve in Appendix \ref{sec:E}.
\end{itemize}



\end{document}